\PassOptionsToPackage{table}{xcolor}
\PassOptionsToPackage{hyphens}{url}

\documentclass[conference,compsoc]{IEEEtran}
\IEEEoverridecommandlockouts 

\usepackage{tikz}
\usetikzlibrary{arrows.meta}
\usepackage{amsmath}
\usepackage{enumitem}
\usepackage{enumerate}
\usepackage{wasysym}
\usepackage{subcaption}
\usepackage{xcolor}
\usepackage{cite}
\usepackage{booktabs}
\usepackage{makecell}
\usepackage{xurl}
\usepackage{hyperref}
\usepackage{adjustbox}
\usepackage{todonotes}
\usepackage{tcolorbox}
\usepackage[framemethod=TikZ]{mdframed}
\usepackage{wrapfig}
\usepackage{cite}

\usepackage{pifont}
\newcommand{\cmark}{\ding{51}} 
\newcommand{\xmark}{\ding{55}} 

\definecolor{policyblue}{HTML}{003366}

\tcbset{
  policybox/.style={
    colback=white,
    colframe=policyblue,
    boxrule=0.8pt,
    arc=2mm,
    left=6pt,right=6pt,top=4pt,bottom=4pt,
    before skip=6pt, after skip=6pt
  }
}

\def\BibTeX{{\rm B\kern-.05em{\sc i\kern-.025em b}\kern-.08em
    T\kern-.1667em\lower.7ex\hbox{E}\kern-.125emX}}

\hypersetup{
    linkcolor=blue,
    citecolor=blue,
    urlcolor=blue,
}

\begin{document}

\title{An Empirical Study and Assessment of EU AI Act Compliance Checkers}



\author{%
\IEEEauthorblockN{%
Zhen Tao\IEEEauthorrefmark{4}\IEEEauthorrefmark{1},
Alize Kahraman\IEEEauthorrefmark{4}\IEEEauthorrefmark{1},
Shidong Pan\IEEEauthorrefmark{2},
Zhenchang Xing\IEEEauthorrefmark{3},\\
Chiara Ullstein\IEEEauthorrefmark{4},
Jens Grossklags\IEEEauthorrefmark{4},
Chunyang Chen\IEEEauthorrefmark{4}}
\IEEEauthorblockA{%
\IEEEauthorrefmark{4}Technical University of Munich, Germany\\
\IEEEauthorrefmark{2}New York University, USA\\
\IEEEauthorrefmark{3}CSIRO’s Data61, Australia\\}
\thanks{\IEEEauthorrefmark{1} Contributed equally to this work.}}

\maketitle

\begin{abstract}
The EU AI Act introduces extensive compliance requirements for organizations that develop, deploy, or integrate AI systems. Many of these requirements are directly relevant to security and privacy, while also addressing closely related issues such as data governance, transparency, accuracy, and robustness. However, stakeholders such as small-to-medium businesses and individual developers often lack the legal expertise required to interpret these obligations and translate them into engineering and governance practices. This disconnect creates challenges for implementing the EU AI Act and may lead to missing safeguards or misdirected development and deployment efforts. To address this, various automated EU AI Act compliance checkers (AIACCs) have emerged, claiming to streamline compliance assessments and provide practical guidance.
In this paper, we present the first empirical study and assessment of AIACCs. We characterize 12 mainstream AIACCs across multiple dimensions, evaluate their legal coverage and alignment, and analyze checker-generated compliance reports for structure, determinacy, and actionability. We find that the quality of AIACCs varies significantly and that they currently can only serve as early-stage orientation tools. Specifically, we observe inconsistent interaction modes and user-friendliness, a tendency to overly simplify or omit key obligations, and a failure to provide determinate, actionable guidance. As a result, reliance on the current generation of AIACCs may foster a false sense of compliance. With our study, we provide a critical baseline of the current AIACC landscape. We further offer design principles for the implementation of more reliable compliance-support tools.

\end{abstract}


\section{Introduction}
\label{sec_intro}
AI systems are increasingly embedded in software products and services in areas such as employment~\cite{surati2026resume}, education~\cite{zhang2023test}, public services~\cite{pan2026creation}, and critical infrastructure~\cite{hoang2026esg}. 
This rapid deployment has brought long-standing concerns about safety, trustworthiness, security, and privacy to the foreground. 
For example, weak data governance processes within AI systems may expose sensitive attributes or propagate biased data~\cite{mehrabi2021survey}. Insecure AI components may create new attack surfaces, such as prompt injection~\cite{greshake2023not, owasp2025promptinjection}. These risks span the full lifecycle of AI systems, from data collection to post-deployment monitoring~\cite{huang2025ai}.

The EU Artificial Intelligence Act (AI Act)~\cite{the_eu_ai_act} is a major regulatory response to these emerging trends, with provisions becoming applicable in stages from February 2025. 
It establishes a risk-based framework for AI systems placed on or used in the EU market, ranging from prohibited practices to high-risk systems, transparency obligations, minimal-risk uses, and general-purpose AI (GPAI) models~\cite{AI_Act}. 
The AI Act's high-risk regime is closely relevant to security and privacy practice and requires software developers and service providers to maintain processes such as risk management, data governance, accuracy, robustness, and cybersecurity~\cite{AI_Act, Nolte_2025}, extending the existing privacy regulatory landscape, such as the General Data Protection Regulation (GDPR)~\cite{GDPR}, into AI-specific contexts.

Meanwhile, the evolving regulatory environment has increasingly become an additional source of organizational overhead for providers that embed AI into their products or services.
For example, Meta declined to release its advanced AI model in the EU~\cite{guardian2024metallama}, citing regulatory uncertainty in a European policy environment shaped by the AI Act.
Further, recent reports have identified compliance-related shortcomings in products developed by major technology companies~\cite{reuters2024aiactchecker}.
The compliance burden is especially significant for small and medium-sized enterprises (SMEs)  and individual developers, who often lack the legal expertise needed to interpret legal terms and satisfy regulatory requirements~\cite{li2022understanding, khandelwal2024unpacking, li2021developers}.
This concern is especially salient in the context of the AI Act, given the lowering of barriers to AI development, the ubiquity of AI systems, as well as the complexity of the regulatory framework~\cite{kilian2025european,langer2025complexities,Nolte_2025}. 


\begin{figure*}[t]
  \centering
  \includegraphics[width=.98\linewidth]{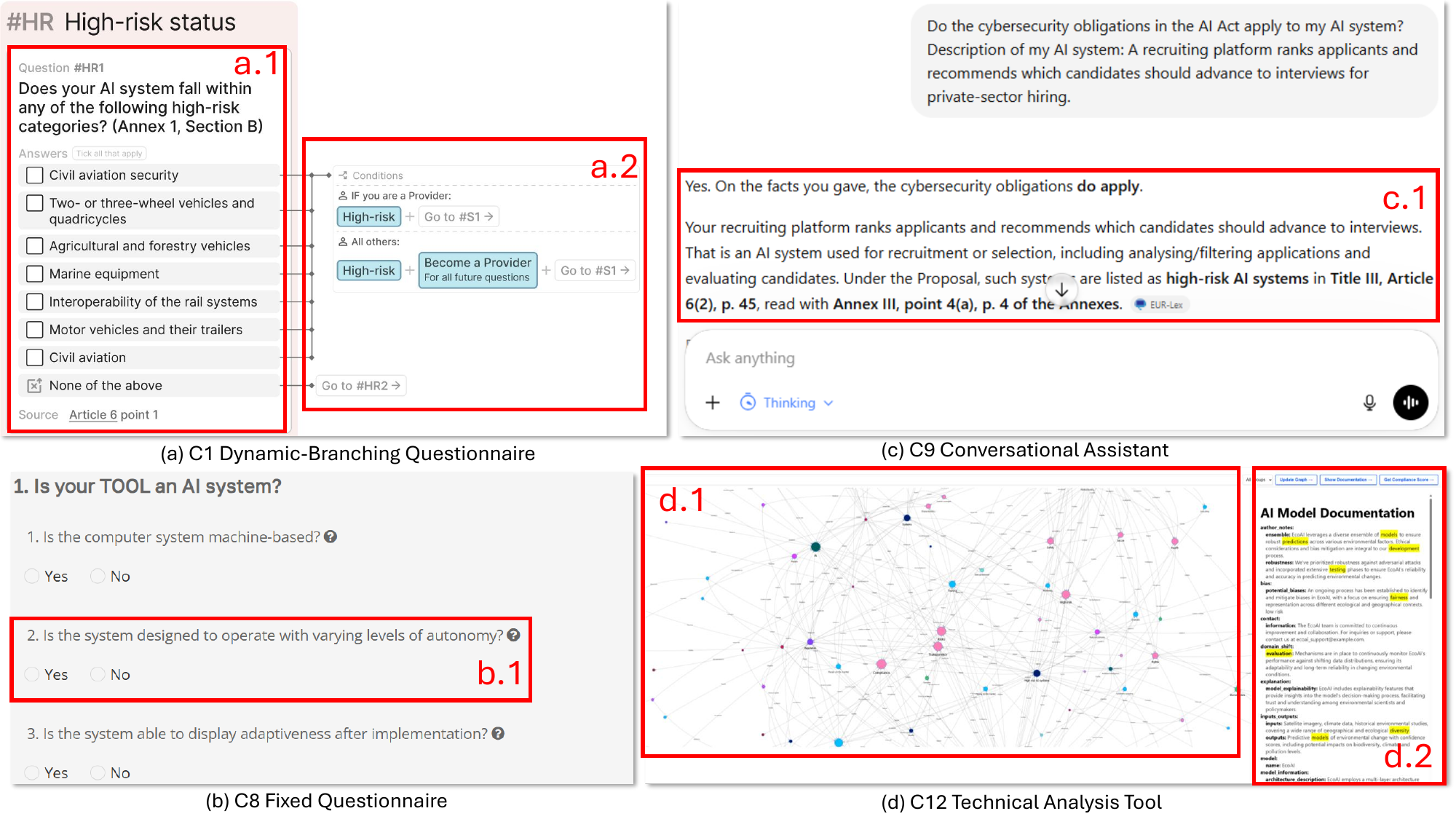}
  \caption{Examples (C1, C8, C9, and C12) of compliance checkers. (a.1) The multiple-choice question in the dynamic-branching questionnaire mode. (a.2) The branching logic of the questionnaire. (b.1) The polar question in the fixed questionnaire mode. (c.1) Conversation regarding cybersecurity obligations in the conversational assistant mode. (d.1) The interactive visualization of AI Act concepts in technical analysis tool mode. (d.2) The assessment of AI model documentation.
  }
  \label{fig_mode}
\end{figure*}

In response, EU AI Act compliance checkers (AIACCs) have emerged as an economic and practical solution for this compliance gap.
Most AIACCs are online tools hosted on websites and accessed through a user interface.
Briefly, a user first needs to describe an AI system, answer a questionnaire, upload materials, or chat with a bot, and AIACCs then perform the compliance checking based on the user inputs. 
Several representative examples are shown in Figure~\ref{fig_mode}.
Such tools \textit{claim} to help developers identify applicable  obligations, understand which safeguards are expected, and provide actionable guidance.
However, their quality, functionality, and other characteristics may vary substantially. 
Unreliable AIACCs could create substantial risks for data governance, personal privacy, and cybersecurity, and these risks may propagate across the broader AI ecosystem.
There is no empirical baseline for characterizing what AIACCs are or for assessing what they actually provide.

To the best of our knowledge, in this paper, we conduct the first systematic characterization of AIACCs.
Inspired by prior studies on compliance-support tools such as automated privacy policy generators~\cite{appg_paper, Sun_2020} and compliance analyzers~\cite{liu2021have}, we conduct an empirical study of AIACCs guided by the following research questions:
\begin{itemize} [leftmargin=*, noitemsep, topsep=3pt]
\item \textbf{RQ1} What AIACCs exist on the market, and what are their differences?
\item \textbf{RQ2} How do AIACCs operationalize compliance checking through their input-output mechanisms, and to what extent do their generated reports provide determinate and actionable guidance?
\item \textbf{RQ3} To what extent do existing AIACCs cover the legal requirements set out in the EU AI Act?
\end{itemize}


Specifically, we assess 12 publicly available AIACCs across three dimensions.
First, we characterize the tool landscape, including interaction mode, user-friendliness, transparency, and recurring design barriers. 
Second, by constructing a corpus of 48 AIACC-generated compliance reports on six representative scenarios, we explore AIACCs' input/output design and analyze the composition, determinacy, and actionability of the created artifacts. 
Third, we evaluate AI Act article coverage and legal alignment of AIACCs. 

Our findings show that the current AIACC landscape is immature. The tools' behavior varies sharply across interaction modes, but they share the goal of lowering the threshold for navigating legal resources by translating legal text into more accessible forms. 
Across modes, we observe shortcomings such as lack of input validation, oversimplified legal modeling, inconsistent and irreproducible results, limited accountability, and readability issues. 
The composition of AIACC-generated compliance reports also varies substantially. Although reports commonly present a risk assessment and applicable obligations, role determination and object determination is often absent, and legal justification is uneven. Consequently, a report may provide a plausible risk label without clearly establishing what contexts support the result.
In addition, AIACC-generated reports often use uncertain or conditional language, and their follow-up recommendations often restate legal obligations rather than specify concrete actors, artifacts, controls, or next steps.
Article coverage is also uneven. AIACCs more consistently cover provisions concerning risk classification, but coverage drops for governance and security-relevant obligations such as data governance, robustness, cybersecurity, and fundamental rights impact assessment.
Lastly, we discuss the security and privacy threat to the broader ecosystem, derive design principles for future AIACC tools, and outline implications and recommendations for stakeholders.

In summary, we make the following contributions:
\begin{itemize} [leftmargin=*, noitemsep, topsep=3pt]
    \item To our knowledge, we offer the first empirical study of publicly available EU AI Act compliance checkers. 
    \item We present a systematic characterization of EU AI Act compliance checkers. We reveal the status quo in terms of their interaction modes, functionalities, and design flaws.
    \item We construct and analyze a corpus of AIACC-generated compliance reports, examining report composition, determinacy, and actionability.
    \item We evaluate legal coverage and alignment against the EU AI Act by testing representative AI usage scenarios, revealing widespread systematic problems.
\end{itemize}

\section{Background}
\label{sec_background}

\subsection{Development and Challenges of the EU AI Act}
The EU AI Act~\cite{the_eu_ai_act} is the first-ever legal framework on AI~\cite{AI_Act}. It aims to ensure that AI systems in the EU market are safe, trustworthy, transparent, and respectful of fundamental rights in cyberspace~\cite{panigutti2023role, plass2022human, ullstein2026proposing, AI_Act}. The Act follows a risk-based structure. Certain unacceptable-risk practices are prohibited. High-risk systems are permitted only under strict obligations. Limited-risk systems are subject mainly to transparency duties, and minimal-risk systems face fewer requirements~\cite{the_eu_ai_act}. The implementation of the AI Act is phased. It entered into force on 1 August 2024, with general-purpose AI (GPAI) provisions becoming applicable on 2 August 2025 and most remaining obligations apply from 2 August 2026~\cite{AI_Act}. Many of its requirements concern security and privacy safeguards around data, system behavior, and accountability. High-risk systems must address obligations such as risk management, data governance, transparency, accuracy, robustness, and cybersecurity~\cite{the_eu_ai_act}. These duties affect systems deployed in sensitive areas that may involve serious privacy and security issues, such as employment, education, critical infrastructure, and public services~\cite{AI_Act}. Therefore, the AI Act shapes the security and privacy work that organizations are expected to perform around them.

However, implementing the AI Act remains challenging~\cite{kilian2025european}, as developers and organizations may struggle with translating legal obligations into engineering and governance practice~\cite{mokander2022conformity,golpayegani2024ai}. This resembles a requirements engineering problem, where abstract legal duties must be interpreted and assigned to responsible actors, and operationalized as technical and organizational controls. Prior work shows that this challenge rises especially when obligations depend on distributed supply chains, evolving standards, or hard-to-test concepts such as human oversight and cybersecurity~\cite{marino2024compliancecardsautomatedeu,kilian2025european,langer2025complexities,Nolte_2025}.

\subsection{Compliance-Support Tools}
With increasing compliance pressure, compliance-support tools have become a popular option for software developers in recent years.
They aim to reduce the cost of interpreting regulations and perform privacy engineering tasks such as privacy notice generation~\cite{yu2015autoppg, liu2021have, appg_paper, zimmeck2021privacyflash, li2024matcha} and compliance checking~\cite{yu2018ppchecker, zimmeck2017automated, andow2019policylint, xiang2023policychecker, wang2022privguard}. 
They make regulatory requirements more accessible to developers who commonly lack legal knowledge~\cite{li2022understanding, li2021developers}. EU AI Act compliance checkers (AIACCs) are an emerging instance of this broader compliance-support tool category and their uptake indicates practical demand. In our selected checkers, C1 had an estimated 344K visits in May 2026 according to SimilarWeb traffic estimates~\cite{similarweb}, suggesting that public-facing AIACCs draw significant attention. This popularity is understandable, but it also raises the stakes of checker quality. If users rely on a low-quality tool to check compliance, incorrect or incomplete assessment result may propagate into downstream stages in the lifecycle, such as documentation, security and privacy safeguard implementation, and deployment decisions. Prior work on automated compliance tools also unveils such systematic issues. For example, automated privacy policy generators can produce policies that are incomplete, inconsistent, or misaligned with app behavior~\cite{appg_paper}, undermining accurate disclosure and user notice. In the AI Act context, such ``butterfly effect'' may also occur where early incorrect assessments could shape later engineering work and leave security or privacy safeguards underdeveloped.

\subsection{Motivation}
The EU AI Act establishes a risk-based framework intended to foster trustworthy AI in cyberspace. 
Security and privacy considerations are deeply embedded in its requirements, including obligations concerning data governance, transparency, accuracy, robustness, and cybersecurity controls. 
We therefore study compliance checkers as a security- and privacy-relevant decision layer between an AI system and an organization's compliance actions, especially security and privacy practices.
We consider a user, such as a developer or an organization without specialized legal expertise, who uses a compliance checker to assess their AI systems. 
The user relies on checker-generated reports to identify and decide what subsequent safeguards to implement. Individuals affected by the assessed AI system, such as its customers, are downstream stakeholders whose safety and privacy may be influenced by the safeguard implementation guided by the compliance checking result.
\section{Methodology}
To characterize AIACCs, we manually collect and identify 12 publicly available AIACCs. 
We construct six representative AI system scenarios as testing inputs. 
Based on this, we evaluate the legal coverage of AIACCs and collect the AIACC-generated compliance reports. 

\subsection{Checker Collection and Selection}

We identify publicly available EU AI Act compliance checkers using the Google search engine. 
We act as a hypothetical software developer who is searching for tools to identify the risk level of their AI system as defined by the AI Act or check the compliance of their AI system. 
We create an initial set of search terms, e.g., \textit{``EU AI Act compliance checker''}, \textit{``AI Act risk classification tool,''} and \textit{``AI Act self-assessment tool''}. In addition, we identify checkers using the search terms in platforms including GitHub~\cite{Github}, GPTs (custom versions of ChatGPT) Store~\cite{openai_gpts} and LinkedIn~\cite{LinkedIn}.

\begin{figure}[t]
  \centering
  \includegraphics[width=.99\linewidth]{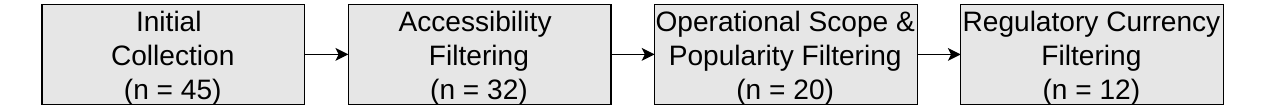}
  \caption{Selection process of compliance checkers.
  }
  \label{fig_checker_selection}
\end{figure}

To evaluate search results, we manually review highly related sources and collect 45 AIACCs as the initial set. We inspected them in September 2025. We filter out checkers based on our exclusion criteria in Figure~\ref{fig_checker_selection}. Specifically, we select AIACCs considering the following dimensions: 
\begin{itemize} [leftmargin=*, noitemsep, topsep=3pt]
\item \textbf{Accessibility}: the checker must be publicly reachable.
\item \textbf{Operational scope}: the checker must perform assessment; purely informational or marketing pages were excluded.  
\item \textbf{Popularity}: we examine the popularity based on number of users, ratings, conversation counts, SimilarWeb traffic estimates~\cite{similarweb}, and GitHub stars. For instance, C1~\cite{tool_aiactchecker2025} reports more than 150,000 monthly users and C9~\cite{tool_chatgpt2025aiactinsight} has over 1,000 user conversations and a rating of 4.2 in GPTs Store.
\item \textbf{Regulatory currency}: the checker must be based on the official Journal version of the EU AI Act rather than early draft versions.
\end{itemize}

We present the 12 selected AIACCs in Table~\ref{tab_checker_list}. ``Provider Sector'' refers to the provider or developer of the checker. For example, C6~\cite{tool_EU_Checker} is provided by the European Commission (public authority) and C9~\cite{tool_chatgpt2025aiactinsight} is developed by a GPT builder (community builder). 
``Registration'' records whether users must create an account or submit identifying information before using the checker. ``Mode'' captures the main interaction model. ``AI-empowered'' indicates whether the checker relies on AI-based functionality rather than rule-based workflows. 
Several checker providers also advertise paid services. 
Payment generally does not unlock additional features within the checker but provides separate offerings such as legal expert consulting and enterprise governance management platforms, we therefore focus on the free checker services in our study.


%
%
\begin{table}[t]
\centering
\caption{Representative scenarios used for AIACC-generated report collection.
}
\label{tab_scenarios}
\resizebox{\linewidth}{!}{%
\begin{tabular}{llll}
\toprule
 & \textbf{Branch} & \textbf{Scenario} & \textbf{Role} \\
\midrule
S1 & Prohibited & Workplace emotion recognition & Deployer \\
S2 & High-risk & Recruitment CV screening & Provider \\
S3 & Transparency & E-commerce chatbot & Deployer \\
S4 & GPAI & GPAI API provider & Provider \\
S5 & Minimal risk & Email spam filter & Deployer \\
S6 & Out of scope & Research-only AI model & Provider \\
\bottomrule
\end{tabular}
}
\end{table}

%
%
\begin{table*}[!t]
\centering
\caption{Compliance checkers selected for our study.}
\resizebox{0.99\linewidth}{!}{%
\begin{tabular}
{cllllc}
\toprule
\# & \textbf{Name} & \textbf{Provider Sector}  & \textbf{Registration}& \textbf{Mode} & \textbf{AI-empowered} \\
\midrule
1 & EU AI Act Compliance Checker~\cite{tool_aiactchecker2025} & Non-profit organization & Not required& Dynamic‑Branching Questionnaire & \xmark \\
2 & TRAIL ML EU AI Act Compliance Checker~\cite{tool_trailml2025checker} & Industry provider & Not required& Dynamic‑Branching Questionnaire & \xmark \\
3 & T\"UV Risk Navigator~\cite{tool_tuev2025risknavigator} & Industry provider  & Not required& Dynamic‑Branching Questionnaire & \xmark  \\
4 & Statworx AI Act Quick Check~\cite{tool_statworx2025quickcheck} & Industry provider & Required& Dynamic‑Branching Questionnaire & \xmark \\
5 & AI Act Conformity Check~\cite{tool_eitdigital2025conformitycheck} & Non-profit organization & Required& Dynamic‑Branching Questionnaire & \xmark \\
6 & EU AI Act Compliance Checker~\cite{tool_EU_Checker} & Public authority & Not required& Dynamic‑Branching Questionnaire & \xmark \\
7 & ARQNXS EU AI Act Compliance Checker~\cite{tool_arqnxs2025homepage} & Industry provider  & Not required& Fixed Questionnaire & \xmark \\
8 & EU AI ACT Self-assessment Questionnaire~\cite{tool_Self_assessment} & Industry provider & Required& Fixed Questionnaire & \xmark \\
9 & EU AI Act Insight GPT~\cite{tool_chatgpt2025aiactinsight} & Community builder & Required & Conversational Assistant & \cmark \\
10 & Whisperly AI Act Checker~\cite{tool_whisperly2025checker} & Industry provider & Required & Conversational Assistant & \cmark \\
11 & COMPL-AI~\cite{tool_COMPL_AI} & Academia   & Not required & Technical Analysis Tool & \xmark \\
12 & AI Compliance Verifier~\cite{tool_gallagher2025aicomplianceverifier} & Academia &Not required& Technical Analysis Tool & \xmark \\
\bottomrule
\end{tabular}%
}
\label{tab_checker_list}
\end{table*}

\subsection{Scenario Construction and Report Collection}
\label{subsec_scenario_construction}
The AI Act follows a risk-based approach, in which obligations depend on whether an AI system falls into unacceptable-risk, high-risk, transparency-risk, minimal-risk, general-purpose AI (GPAI), or out-of-scope categories~\cite{AI_Act, the_eu_ai_act}. To evaluate legal alignment and report content under this structure, we construct six representative scenarios (S1--S6), each targeting a distinct regulatory branch. We summarize the scenarios in Table~\ref{tab_scenarios}. Detailed structured fact sheets are released as artifacts.
We construct the scenarios to cover both legally explicit cases and common real-world uses. S1 targets workplace emotion recognition, an Article 5 prohibited practice discussed in official AI Act materials~\cite{AI_Act}. S2 captures recruitment and candidate-ranking systems, a canonical high-risk employment use case mentioned in the text of AI Act~\cite{the_eu_ai_act}. S3 covers a customer-service chatbot, a common transparency-risk case under Article 50~\cite{AI_Act}. S4 targets GPAI providers~\cite{ec2025gpaiguidelines, ec2025gpaicode}. S5 defines an email spam filter, which official materials describe as minimal or no risk~\cite{AI_Act}. S6 tests the edge case exclusion for systems developed and tested solely for scientific research without being placed on the market or put into service~\cite{the_eu_ai_act}. For report content analysis, we focus on questionnaire-based AIACCs (C1-C8) as they are the dominant mode and consistently return report-like outputs. We collect one report per checker-scenario pair, yielding 48 reports from eight AIACCs and six scenarios. Specifically, we translate the six scenario fact sheets into answers to each checker's questions. We examine the paths reached by these scenarios and collect their resulting assessment report. We release the report corpus at:
~\url{https://github.com/ZhenTAO3059/AIACC}.












\subsection{Annotation Procedure}

Across 12 AIACCs, we manually examine the interaction mechanism, requested inputs, available outputs, and disclosed checking logic. We annotate characteristics and input/output features using the annotation criteria presented in Section ~\ref{app_def_c} and ~\ref{app_def_io} in the Appendix, and examine EU AI Act article coverage within the scope specified in Section~\ref{sec_rq2}. 
Two annotators independently apply these criteria and then resolve disagreements through discussion and reinspection. After discussion, the annotators refined the criteria and recoded the affected cases. One example of disagreement is whether exposing reasoning explanations can be considered as transparency of checking logic. The annotators reached a consensus to label this as \LEFTcircle{}. As the annotation task was lightweight and the number of annotations was small, we did not compute an inter-annotator agreement statistic but report the consensus annotations in the following sections.
One annotator has more than three years of research experience in privacy and regulatory compliance analysis, and the other works at a well-known company (10,000+ employees) testing, inspecting, and certifying technical systems and facilities. 
\section{Status Quo of the Checkers (RQ1)}
\label{sec_rq1}

In this section, we present the status quo of the emerging ecosystem of EU AI Act compliance checkers. We analyze the characteristics and design flaws of 12 selected AIACCs.
Among them, AIACC providers span five groups. 

\subsection{Modes}

Based on their underlying working mechanisms and types of user interaction, we categorize AIACCs into four modes: (a) dynamic branching questionnaire, (b) fixed questionnaire, (c) conversational assistant, and (d) technical analysis tool, as shown in Figure~\ref{fig_mode}.

\textbf{Dynamic Branching Questionnaires} require users to complete a questionnaire incorporating dynamic logic to describe their AI system, and checks its compliance based on the provided answers. The underlying logic operates by dynamically tailoring the checking path. One example is presented in Figure~\ref{fig_mode}(a), that C1 selects the next path based on the dynamic input from users. \textbf{Fixed Questionnaire Tools} present a linear, non-adaptive series of questions. As shown in Figure~\ref{fig_mode}(b), C8 poses yes-no polar questions. \textbf{Conversational Assistants} rely on chatbot-style interaction powered by large language models (LLMs) to guide users through compliance questions or respond to users' inquiries in natural language. These systems may support both user-led exploration and tool-led assessment. C9 is a custom GPT available in GPTs Store engaging in conversations with users regarding compliance with the EU AI Act. Differently, C10's logic is largely scripted and leads the assessment in a chatbot style. Thus, conversational mode does not necessarily imply free-form reasoning or inference. This mode is the only mode that integrates AI-based functionality, e.g., support chatbot-style checking. \textbf{Technical Analysis Tools} rely on analyzing materials, such as AI model outputs and documentation, to conduct compliance checks. For example, C11 maps AI Act principles to a model's benchmark results and returns its compliance scores. C12 transforms the legal text of the Act into a knowledge graph and derives compliance checklists, and compares them against model documentation to identify missing items and generate recommendations. Despite their differing forms, AIACCs across various modes share the goal of simplifying complex legal text into easily understandable questions or other forms of representation.

Branching questionnaires approximate the AI Act's legal structure and fixed questionnaires privilege speed. Conversational assistants stand out for accessibility and user-friendliness, and technical tools focus on inspectability and artifact-level analysis.
These differences suggest an inherent trade-off between legal fidelity, user-friendliness, transparency, and evidentiary rigor. 
The interaction mode of an AIACC not only shapes the experience but also influences the depth, reliability, and actionability of the compliance guidance it provides.


\subsection{Characterization}

%
%
\begin{table}[t]
\centering
\caption{The characterization of 12 AIACCs. \CIRCLE: high-level support; \LEFTcircle: intermediate support; \Circle: low-level support. ``DBQ'' refers to dynamic branching questionnaire; ``FQ'' refers to fixed questionnaire; ``CA'' refers to conversational assistant; ``TAT'' refers to technical analysis tool.}
\label{tab_char}
\scriptsize
\setlength{\tabcolsep}{2.2pt}
\renewcommand{\arraystretch}{1.08}
\resizebox{\linewidth}{!}{%
\begin{tabular}{l|cccccc|cc|cc|cc}
\hline
\rowcolor{lightgray!15}
\textbf{Mode} & \multicolumn{6}{c|}{\textbf{DBQ}} & \multicolumn{2}{c|}{\textbf{FQ}} & \multicolumn{2}{c|}{\textbf{CA}} & \multicolumn{2}{c}{\textbf{TAT}} \\

\rowcolor{lightgray!85}
\textbf{\# of AIACC} & \textbf{1} & \textbf{2} & \textbf{3} & \textbf{4} & \textbf{5} & \textbf{6} & \textbf{7} & \textbf{8} & \textbf{9} & \textbf{10} & \textbf{11} & \textbf{12} \\
\hline

\rowcolor{lightgray!15}
Risk Classification & \CIRCLE & \CIRCLE & \CIRCLE & \CIRCLE & \CIRCLE & \CIRCLE & \Circle & \CIRCLE & \CIRCLE & \CIRCLE & \Circle & \Circle \\

\rowcolor{lightgray!50}
Article Mapping & \CIRCLE & \LEFTcircle & \CIRCLE & \CIRCLE & \CIRCLE & \CIRCLE & \LEFTcircle & \Circle & \LEFTcircle & \LEFTcircle & \CIRCLE & \LEFTcircle \\


\rowcolor{lightgray!50}
User Guidance & \CIRCLE & \CIRCLE & \CIRCLE & \LEFTcircle & \LEFTcircle & \CIRCLE & \Circle & \LEFTcircle & \CIRCLE & \CIRCLE & \Circle & \Circle \\

\rowcolor{lightgray!15}
Examples & \Circle & \Circle & \Circle & \Circle & \Circle & \LEFTcircle & \CIRCLE & \CIRCLE & \LEFTcircle & \Circle & \Circle & \Circle \\

\rowcolor{lightgray!50}
Interactive Q\&A & \Circle & \Circle & \Circle & \Circle & \Circle & \Circle & \Circle & \Circle & \CIRCLE & \Circle & \Circle & \Circle \\

\rowcolor{lightgray!15}
Data Handling & \CIRCLE & \CIRCLE & \CIRCLE & \CIRCLE & \CIRCLE & \CIRCLE & \CIRCLE & \CIRCLE & \Circle & \CIRCLE & \CIRCLE & \CIRCLE \\

\rowcolor{lightgray!50}
Checking Logic & \CIRCLE & \Circle & \LEFTcircle & \Circle & \Circle & \LEFTcircle & \LEFTcircle & \Circle & \LEFTcircle & \LEFTcircle & \CIRCLE & \CIRCLE \\

\rowcolor{lightgray!15}
Result Export & \CIRCLE & \Circle & \CIRCLE & \Circle & \CIRCLE & \CIRCLE & \CIRCLE & \CIRCLE & \Circle & \Circle & \CIRCLE & \Circle \\

\hline
\end{tabular}
}
\end{table}

We assess 12 AIACCs along various dimensions. The dimensions are separated into 3 groups: (i) functionality (1-2); (ii) user support (3-5); (iii) transparency and accountability (6-8). The results are presented in Table~\ref{tab_char}. 
We present the definition of each item in the Appendix~\ref{app_def_c}.

\textbf{Functionality.} Risk classification is a common capability across AIACCs. This pattern aligns with the structure of the AI Act, which is organized around risk levels of AI systems. Classification provides an intuitive first answer for users. 
Several AIACCs can return a risk level while providing weaker article mapping or limited explanation of why a particular provision applies. 
Questionnaires generally provide strong legal mapping as their control flow resembles the Act's conditional structure. Conversational assistants mention articles, but their mappings are less stable and depend on prompt wording. Differently, technical analysis tools do not necessarily cover risk classification and article mapping as they focus more on analyzing specific materials.

\textbf{User Support.} The main target audience of AIACCs is expected to be organizations and developers with little or no legal knowledge about the compliance with the AI Act. Therefore it is necessary to provide sufficient user support. Especially in questionnaire mode, incomplete or incorrect user input can affect the quality of the output. We consider user support as the following: whether the checker provides user guidance, whether it provides specific scenario examples, and whether there are interactive Q\&A.

\textbf{Transparency and Accountability.}
We refer transparency to whether AIACCs disclose how user inputs are interpreted, how legal provisions are reflected, and how checking results are generated (``checking logic'').  ``Data handling'' indicates whether AIACCs are transparent about the collection and processing of potentially sensitive data during the checking (e.g., through privacy policies). Technical analysis tools are open-source projects and provide more inspectable logic, but they shift the burden to technically sophisticated users. The opacity regarding the checking mechanisms may affect users' trust in the results of compliance checks. Compliance checking also produces artifacts that users may reuse internally for documentation and decision-making. However, ``Result export'' is inconsistent. Some checkers produce compliance reports that can be downloaded, while others display results transiently.

The characterization reveals both good practices and weak practices in the current AIACC ecosystem. C1, C3, and C6 are overall the strongest, covering the core functionality of risk classification and article mapping, provide user guidance, disclose data handling, and offer reusable results. C1 further provides checking-logic transparency. Many AIACCs decently cover some important dimensions, but their support is uneven. C2, C7, and C12 merely cover selected dimensions well. Above all, users should carefully choose AIACCs that best meet their requirements.



\begin{figure}[t]
  \centering
  \includegraphics[width=.99\linewidth]{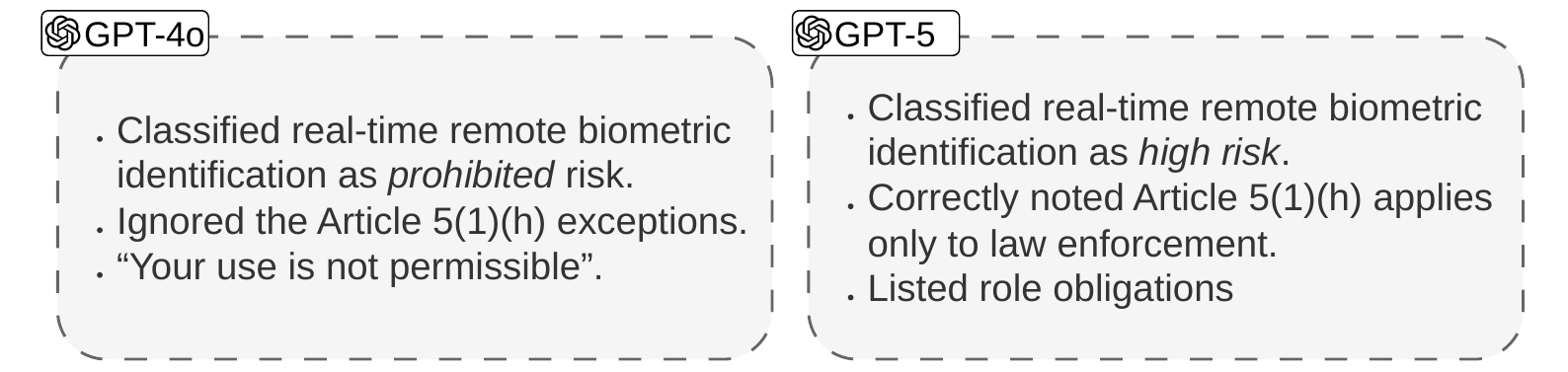}
  \caption{Different results using different base models under identical prompt. GPT-4o produced incorrect result (left box) while GPT-5 produced correct result (right box).
  }
  \label{fig_model_incon}
\end{figure}

\begin{figure}[t]
  \centering
  \includegraphics[width=.99\linewidth]{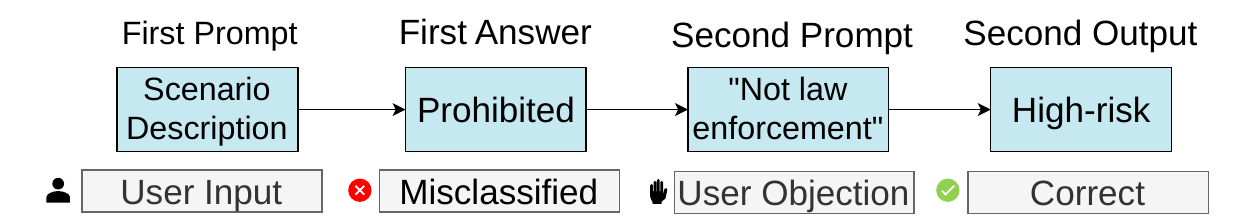}
  \caption{A compliance checking conversation flow indicating inconsistency and user-pleasing behavior. The checker first produced a misclassified \emph{prohibited} answer, then changed to \emph{high-risk} after user objection.
  }
  \label{fig_flow_incon}
\end{figure}



\begin{tcolorbox}[colback=gray!10, colframe=gray!60!black, boxrule=0.5pt, arc=3pt]
\textbf{Findings to RQ1:} AIACCs are designed to lower the legal entry barrier. They translate convoluted legal text into more accessible forms. AIACCs have various characteristics influenced by the differing focuses of providers. This variety provides users with choices, while also reflecting a trade-off between utility, user-friendliness, and transparency. 
\end{tcolorbox}


%
%
\begin{figure}[t]
    \centering
    \begin{subfigure}[t]{\linewidth}
        \centering
        \includegraphics[width=0.85\linewidth]{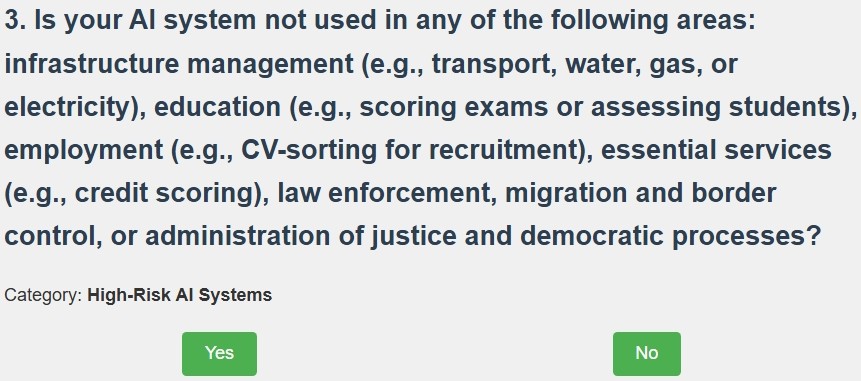}
        \caption{C7: Confusing double-negative phrasing of a yes/no question.} 
    \end{subfigure}
    
    \vspace{0.75em}
    
    \begin{subfigure}[t]{\linewidth}
        \centering
        \includegraphics[width=0.85\linewidth]{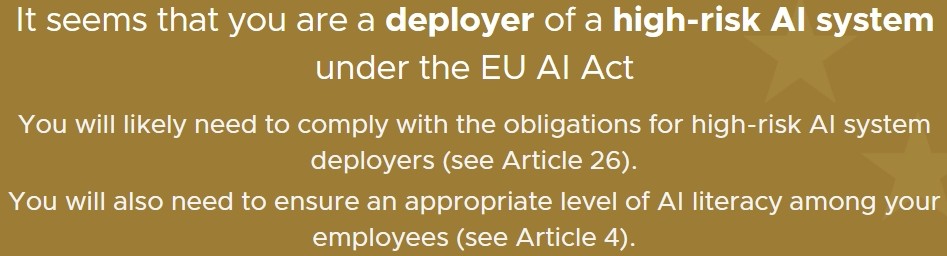}
        \caption{C2: Ambiguous output using inconsistent confidence levels (``\textit{will likely need to...}'' and ``\textit{will also need to...}'').} \label{fig:readabilityweaknessp2}
    \end{subfigure}
    \caption{Examples of readability issues in AIACCs.}
    \label{fig_readability}
\end{figure}

%
%







%
%
\begin{table}[t]
\centering
\caption{The breakdown of required inputs and generated outputs of AIACCs.}
\label{tab_inputoutput}
\scriptsize
\setlength{\tabcolsep}{2.2pt}
\renewcommand{\arraystretch}{1.08}
\resizebox{\linewidth}{!}{%
\begin{tabular}{l|cccccc|cc|cc|cc}
\hline
\rowcolor{lightgray!15}
\textbf{Mode} & \multicolumn{6}{c|}{\textbf{DBQ}} & \multicolumn{2}{c|}{\textbf{FQ}} & \multicolumn{2}{c|}{\textbf{CA}} & \multicolumn{2}{c}{\textbf{TAT}} \\

\rowcolor{lightgray!85}
\textbf{\# of AIACC} & \textbf{1} & \textbf{2} & \textbf{3} & \textbf{4} & \textbf{5} & \textbf{6} & \textbf{7} & \textbf{8} & \textbf{9} & \textbf{10} & \textbf{11} & \textbf{12} \\
\hline

\rowcolor{lightgray!50}
\multicolumn{13}{l}{\textbf{Input}} \\

\rowcolor{lightgray!15}
System Description & \Circle & \Circle & \Circle & \Circle & \CIRCLE & \Circle & \Circle & \Circle & \LEFTcircle & \Circle & \CIRCLE & \CIRCLE \\

\rowcolor{lightgray!50}
User's Contact Information & \Circle & \LEFTcircle & \Circle & \CIRCLE & \CIRCLE & \Circle & \Circle & \CIRCLE & \Circle & \LEFTcircle & \Circle & \Circle \\

\rowcolor{lightgray!15}
AI Model or System & \CIRCLE & \CIRCLE & \CIRCLE & \CIRCLE & \Circle & \CIRCLE & \Circle & \CIRCLE & \LEFTcircle & \CIRCLE & \Circle & \Circle \\

\rowcolor{lightgray!50}
Domain of Application & \CIRCLE & \CIRCLE & \CIRCLE & \CIRCLE & \CIRCLE & \CIRCLE & \CIRCLE & \CIRCLE & \CIRCLE & \CIRCLE & \Circle & \Circle \\

\rowcolor{lightgray!15}
Intended Purpose & \CIRCLE & \CIRCLE & \CIRCLE & \CIRCLE & \CIRCLE & \CIRCLE & \CIRCLE & \CIRCLE & \CIRCLE & \CIRCLE & \Circle & \Circle \\

\rowcolor{lightgray!50}
Interaction Type & \CIRCLE & \CIRCLE & \CIRCLE & \CIRCLE & \Circle & \CIRCLE & \CIRCLE & \Circle & \CIRCLE & \CIRCLE & \Circle & \Circle \\

\rowcolor{lightgray!15}
Biometric/Sensitive Functions & \CIRCLE & \CIRCLE & \CIRCLE & \CIRCLE & \CIRCLE & \CIRCLE & \CIRCLE & \CIRCLE & \CIRCLE & \CIRCLE & \Circle & \Circle \\

\rowcolor{lightgray!50}
User Role & \CIRCLE & \CIRCLE & \CIRCLE & \Circle & \Circle & \CIRCLE & \Circle & \CIRCLE & \CIRCLE & \CIRCLE & \Circle & \Circle \\

\rowcolor{lightgray!15}
Geographic Scope & \CIRCLE & \CIRCLE & \CIRCLE & \CIRCLE & \CIRCLE & \CIRCLE & \Circle & \CIRCLE & \CIRCLE & \CIRCLE & \Circle & \Circle \\

\rowcolor{lightgray!50}
Documentation & \Circle & \Circle & \Circle & \Circle & \Circle & \Circle & \Circle & \Circle & \CIRCLE & \Circle & \CIRCLE & \CIRCLE \\

\hline
\rowcolor{lightgray!50}
\multicolumn{13}{l}{\textbf{Output}} \\

\rowcolor{lightgray!15}
Resulting Report & \CIRCLE & \CIRCLE & \CIRCLE & \CIRCLE & \CIRCLE & \CIRCLE & \CIRCLE & \CIRCLE & \LEFTcircle & \CIRCLE & \CIRCLE & \CIRCLE \\

\rowcolor{lightgray!15}
Benchmark Score & \Circle & \Circle & \Circle & \Circle & \Circle & \Circle & \Circle & \Circle & \Circle & \Circle & \CIRCLE & \Circle \\

\rowcolor{lightgray!15}
Visualization & \Circle & \Circle & \Circle & \Circle & \Circle & \Circle & \Circle & \Circle & \Circle & \Circle & \Circle & \CIRCLE \\









\hline
\end{tabular}
}
\end{table}

%
%

\section{Input-Output Mechanisms (RQ2)}
\label{sec_rq3_report}

AIACCs require user input materials and produce compliance checking results that may shape subsequent organizational compliance decisions.
In this section, we first characterize the input and output artifacts across all 12 AIACCs, and then analyze the composition, determinacy, and actionability of their reports.

\subsection{Breakdown of Inputs and Outputs}
An AIACC's compliance conclusion is shaped not only by its checking logic, but also by the information it elicits and the form in which it communicates the result. If a checker omits legally relevant facts or relies only on unsupported self-claims, its output may rest on an incomplete evidentiary basis, and users may be unable to verify or reuse it in subsequent compliance work. Breaking down the input and output artifacts therefore reveals what each checker can know about an AI system and what kind of compliance artifact it ultimately provides. 
As shown in Table~\ref{tab_inputoutput}, AIACCs differ in what they require from users. Across questionnaire and conversational modes, the common input categories are application domain, intended purpose, interaction type, sensitive functions, geographic scope, and whether the assessed object is an AI model or AI system. These categories map to the core legal routing questions of the AI Act about what the system is, where and why it is used, and whether it involves practices that may trigger obligations. User contact information is administrative rather than legally diagnostic. Conversational assistants accept free-text input, which offers flexibility but shifts the burden to users to describe their AI systems sufficiently and accurately. Technical analysis tools accept documentation, such as benchmarking outputs or model documentation, as input to perform evidence-based assessment. This creates a higher barrier for non-technical users, but it also enables more evidence-based checking.

\begin{figure}[t]
  \centering
  \includegraphics[width=.99\linewidth]{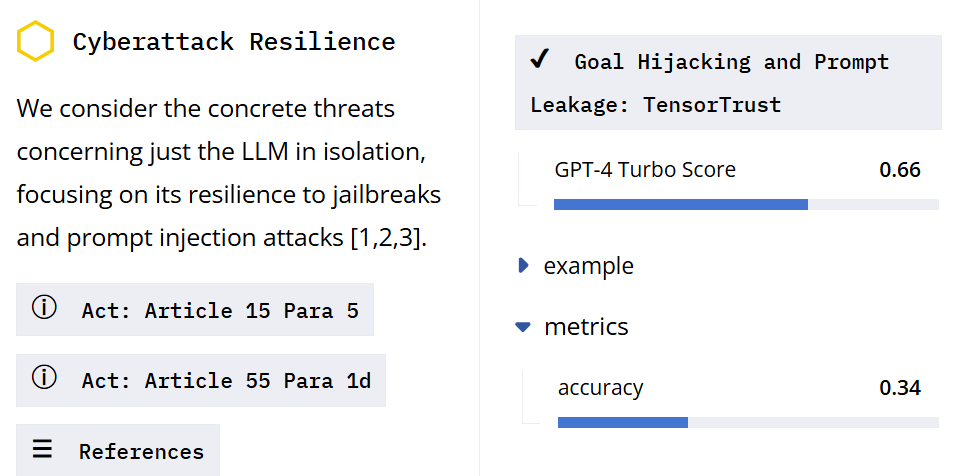}
  \caption{Benchmark score generated by C11 for the \textit{Cyberattack Resilience} requirement in AI Act Articles 15 \& 55. Scores are calculated based on the benchmarking results.
  }
  \label{fig_benchmark_score}
\end{figure}

The outputs of AIACCs also vary by mode. The dominant output is a structured textual report, such as a downloadable questionnaire result or a structured assessment of uploaded artifacts. Reports from C7 and C12 also include compliance scores, which can be confusing because the basis for the number is not always clear. A score may be computed from user self-reported questionnaire answers or from a documentation-matching procedure, without necessarily reflecting whether the AI system has passed any quantitative test. Differently, the benchmark score produced by C11 in Figure~\ref{fig_benchmark_score} is tied to benchmark results for specific obligations, such as cyberattack resilience. C12 can produce visualizations of the AI Act text and uploaded model documentation as knowledge graphs, as shown in Figure~\ref{fig_mode}(d). These outputs are more inspectable, but they require users to interpret technical artifacts rather than simply read a compliance label.



\begin{table}
\caption{Basic statistics of AIACC-generated reports.}
\label{table_report_basic}
\centering
\resizebox{\linewidth}{!}{%
\begin{tabular}{lrrrrrrrr}
\toprule
\textbf{Metric} & \textbf{C1} & \textbf{C2} & \textbf{C3} & \textbf{C4} & \textbf{C5} & \textbf{C6} & \textbf{C7} & \textbf{C8} \\
\midrule
\textbf{Avg. \#Words} & 181.0 & 68.2 & 119.7 & 46.3 & 303.0 & 335.2 & 1023.3 & 324.8 \\
\textbf{Avg. \#Sent.} & 11.7 & 4.2 & 12.7 & 3.3 & 21.0 & 16.7 & 47.7 & 31.5 \\
\bottomrule
\end{tabular}
}%
\end{table}



\subsection{Report Composition}
\label{subsec_composition}
The AIACC-generated report is the primary artifact through which users consume an AIACC's assessment and decide what to do next. We consider the report composition therefore indicates the checker's practical effectiveness. A report must communicate not only a conclusion, but also the regulated object and actor, applicable obligations, supporting legal rationale, and sufficiently specific guidance for follow-up action. Examining what these reports contain and how their content is organized helps reveal whether they convey scenario-specific compliance result or merely present a plausible-looking result.
For report analysis, we focus on questionnaire-based AIACCs (C1-C8), which are the dominant mode. Although conversational assistants and technical analysis tools can also produce outputs, their artifacts are not directly comparable in report corpus. Questionnaire-based tools, by contrast, consistently return report-like outputs for the six scenarios and therefore allow controlled comparison across checker and scenario. 

We display the basic statistics of 48 AIACC-generated reports on 6 representative scenarios in Table~\ref{table_report_basic}. Report length varies substantially across the studied questionnaire-based checkers. C4 produces the shortest reports, averaging 46.3 words and 3.3 sentences, whereas C7 produces the longest, averaging 1,023.3 words and 47.7 sentences. This wide range indicates that users may receive information from a brief classification result to an extensive multi-section assessment. Length alone, however, does not establish whether a report contains the necessary elements.

\begin{figure}[t]
  \centering
  \includegraphics[width=.98\linewidth]{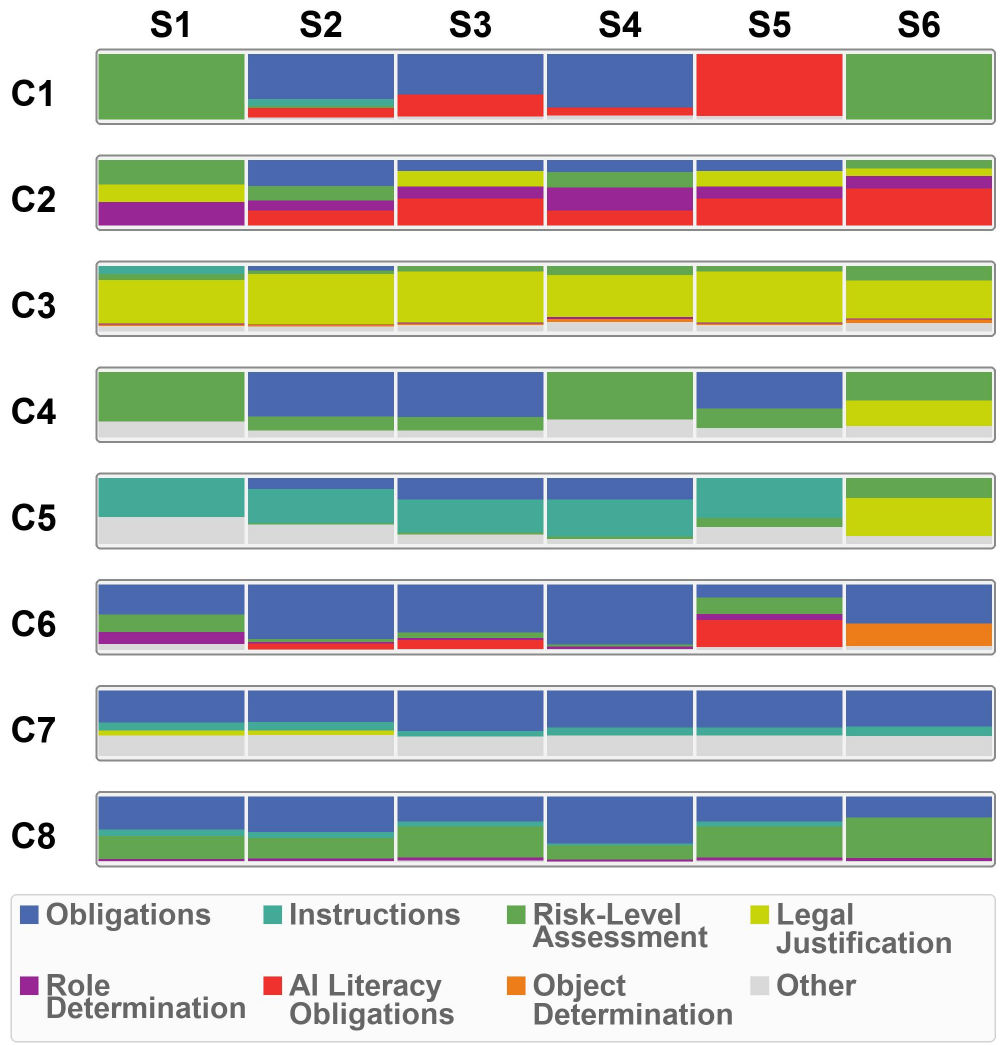}
  \caption{Category composition of checker-generated reports by checker and scenario.
  }
  \label{fig_report_category}
\end{figure}

We display the identified elements in reports in Table~\ref{table_structure_boilerplate} and Figure~\ref{fig_report_category}. Obligations and risk-level assessment are the most frequently covered categories, each appearing in reports from all eight checkers. Instructions and legal justification tie for the next-highest coverage, each appearing in five checkers. In contrast, role determination appears in only four checkers, AI literacy obligations in three, and object determination in two. At the checker level, C3 covers seven categories, the widest range among the studied questionnaire-based checkers, whereas C4, C5 and C7 cover only four. Coverage breadth does not imply balanced content. As shown in Figure~\ref{fig_report_category}, C6 is the most concentrated checker, with obligations accounting for 75.7\% of its report text. C3 is similarly uneven, with legal justification accounting for 70.8\% of its text, while obligations comprise 66.6\% of C1's text. Such concentration can leave users with extensive material about one aspect of compliance but insufficient support for other aspects.

\begin{table}[t]
\caption{Composition of checker-generated reports. *Disclaimers may appear outside of the report (e.g., within text on the webpage). We mark ``\cmark'' whenever disclaimers are found, regardless of the location.}
\label{table_structure_boilerplate}
\centering
\resizebox{\linewidth}{!}{%
\begin{tabular}{lcccccccc}
\toprule
\textbf{Section Category} & \textbf{C1} & \textbf{C2} & \textbf{C3} & \textbf{C4} & \textbf{C5} & \textbf{C6} & \textbf{C7} & \textbf{C8} \\
\midrule
Role determination          & \xmark & \cmark  & \cmark  & \xmark & \xmark & \cmark  & \xmark & \cmark  \\
Risk-level assessment       & \cmark   & \cmark  & \cmark  & \cmark   & \cmark   & \cmark  & \cmark   & \cmark  \\
Object determination   & \xmark & \xmark & \cmark & \xmark & \xmark & \cmark  & \xmark & \xmark \\
Legal justification         & \xmark & \cmark  & \cmark  & \cmark   & \cmark   & \xmark & \cmark   & \xmark \\
Obligations                 & \cmark   & \cmark  & \cmark  & \cmark   & \cmark   & \cmark  & \cmark   & \cmark  \\
AI literacy obligations     & \cmark   & \cmark  & \xmark & \xmark & \xmark & \cmark  & \xmark & \xmark \\
Disclaimer*                  & \cmark & \cmark & \cmark & \cmark & \xmark & \cmark & \xmark & \cmark \\
Instructions                & \cmark   & \xmark & \cmark  & \xmark & \cmark   & \xmark & \cmark   & \cmark  \\
\bottomrule
\end{tabular}
}%
\end{table}

\subsection{Determinacy and Actionability}


We examine two user-facing properties of AIACC-generated reports. \emph{Determinacy} describes whether a report states a conclusion directly or qualifies it with uncertain, conditional, or weakened language. \emph{Actionability} concerns whether directive language is used to communicate steps that users can perform. We use the following metrics as screening indicators of these properties rather than complete measures of report quality.

\textbf{Lexical Hedge Detection.} We use hedge cues as a proxy for low determinacy. Hedging marks whether a statement is presented as possible, conditional, uncertain, or epistemically weakened. Following the BioScope hedge annotation framework~\cite{szarvas2008bioscope} and the CoNLL-2010 hedge-cue task~\cite{farkas2010conll}, we implement a cue-based detector using 213 hedge cues from CoNLL-2010, including auxiliaries, speculative verbs, adjectives, adverbs, and conjunctions. For each report, we compute the Hedge Sentence Rate, which is the proportion of analyzed sentences containing at least one hedge cue. A higher rate indicates that the report relies more heavily on uncertain, conditional, or epistemically softened language, and is therefore treated as less determinate in our analysis. For example, C1's GPAI report states, ``\textit{If so, obligations on high risk AI systems may apply directly or indirectly under Recital 85.}'' This has weaker determinacy because it conditions the conclusion without identifying how the user can resolve the uncertainty.

\begin{figure}[t]
  \centering
  \includegraphics[width=.9\linewidth]{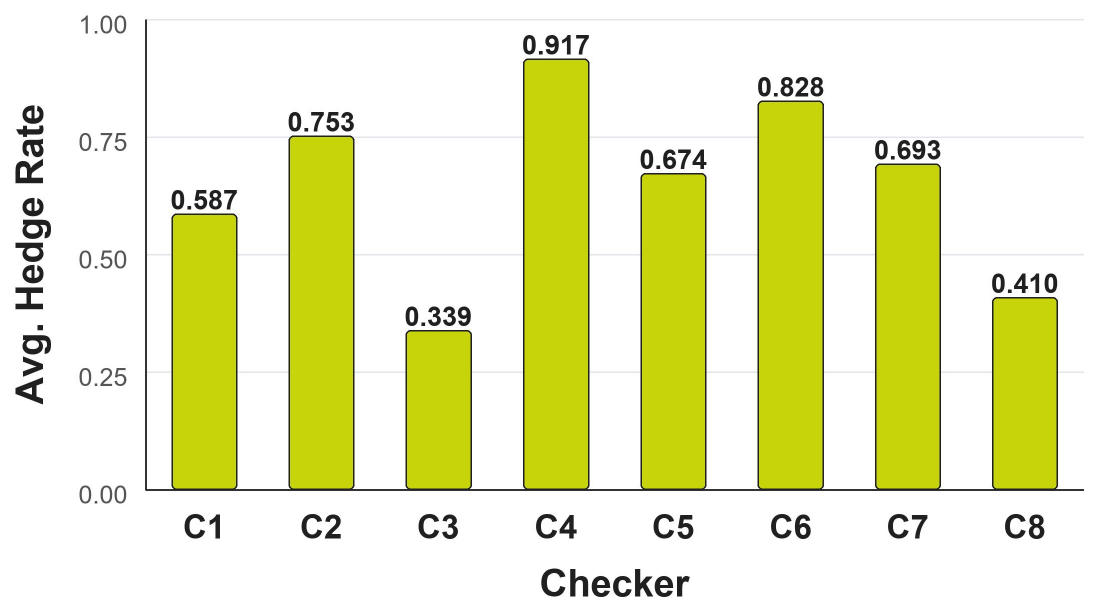}
  \caption{
  Average Hedge Sentence Rate by checker across the six scenarios. Higher values indicate greater hedge-cue prevalence and weaker linguistic determinacy.
  }
  \label{fig_hedge}
\end{figure}

As shown in Figure~\ref{fig_hedge}, there is substantial hedging across reports, with an average hedge sentence rate of 0.650. This is higher than rates reported in BioScope subcorpora (between 0.134 and 0.223)~\cite{szarvas2008bioscope} and CoNLL-2010 datasets (between 0.16 and 0.23)~\cite{farkas2010conll}. Please note that these values provide scale rather than a matched baseline because scientific prose and compliance reports serve different purposes. Our detector should be interpreted as a screening measure of hedge-cue prevalence. Some hedging is appropriate because legal compliance can depend on facts not available to the checker. However, reports often fail to distinguish uncertainty caused by missing facts from qualification required by the law. A better report would state the current conclusion, identify the unresolved fact, and explain how the answer would change once that fact is supplied.

\textbf{Keyword Loss of Specificity.} We assess one aspect of actionability using Keyword Loss of Specificity, which measures whether directive keywords such as ``must,'' ``should,'' and ``need,'' pinpoint actionable advice or occur in other information~\cite{rostami2024qualitative}. A report may contain many compliance verbs while still failing to tell the user what to do next. For each report, we divide the number of keyword-containing sentences not labeled as instructions by all keyword-containing sentences. Thus, 0 means every matched keyword occurs in a sentence labeled as an instruction, while 1 means none does. For example, C1's S3 report states, ``\textit{You need to follow these transparency obligations under Article 50, point 2}''. This sentence is an obligation statement rather than an instruction specifying a procedure. By contrast, C5's S1 report states, ``\textit{According to the current EU AI Act legislation your system falls under the unacceptable risk category, therefore you need to reconsider your usage either by covering a different application area or limiting the usage of your application to the cases...}'' This is labeled as an instruction because it proposes a change to the intended use.

\begin{figure}[t]
  \centering
  \includegraphics[width=.9\linewidth]{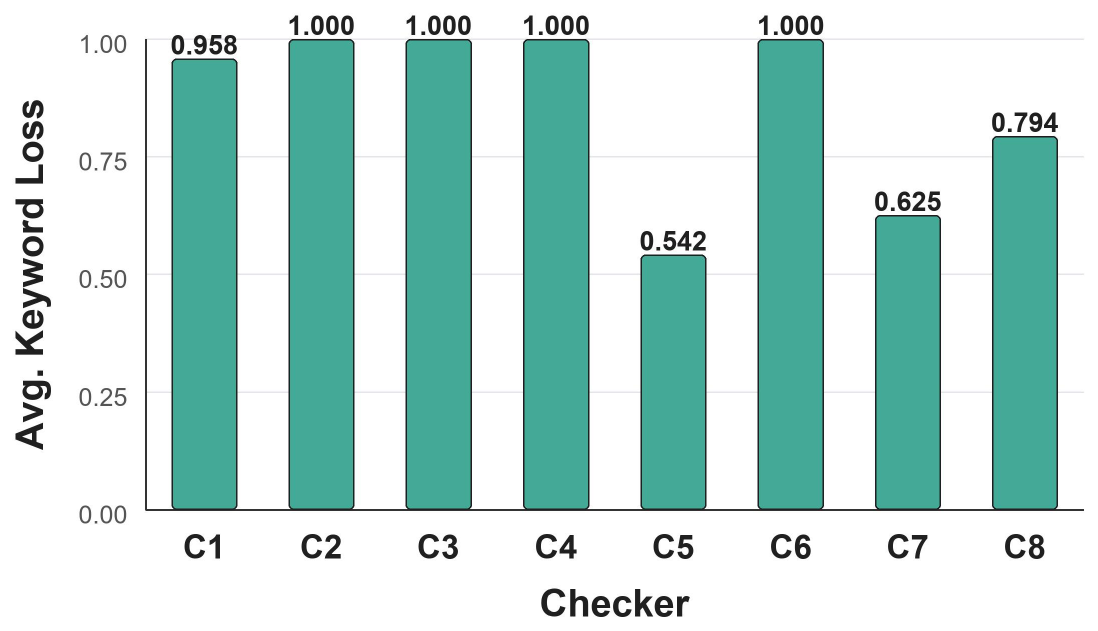}
  \caption{
  Average Keyword Loss of Specificity by checker across generated reports. Lower values indicate that directive keywords more often correspond to direct instructions.
  }
  \label{fig_keyword_loss}
\end{figure}

We present the result in our corpus in Figure~\ref{fig_keyword_loss}. The mean of report-level losses is 0.840, while the pooled loss is 0.764 (168 of 220 keyword-containing sentences). These values show that directive vocabulary usually occurs outside sentences labeled as instructions. Rostami and Karlsson~\cite{rostami2024qualitative} reported a pooled Keyword Loss of Specificity of 0.669 for 15 information-security policies and characterize this result as poor. If obligation statements and AI literacy obligations are also counted as actionable, the mean report-level loss drops from 0.840 to 0.326, and the pooled loss drops from 0.764 to 0.150. This shows that many matched sentences identify duties without giving procedural steps. 
This reflects the intended purpose of AIACCs. For example, C3 claims to ``Classify Your Risk According to the EU AI Act'', and identifying relevant duties may fulfill the stated aim.

Taken together, no checker performs consistently well on both determinacy and actionability. C3 and C8 perform best in terms of determinacy, with the lowest average Hedge Sentence Rates (0.339 and 0.410). C1 is in the middle (0.587), while C4 and C6 perform worst on this dimension (0.917 and 0.828), meaning that their reports are dominated by hedge-cue language. C5 and C7 perform best on Keyword Loss of Specificity (0.542 and 0.625), indicating that their directive keywords more often correspond to direct instructions. Other checkers perform poorly on this metric, meaning their directive language usually appears outside operational guidance. This divergence suggests that a report can sound determinate without being actionable, or contain actionable phrasing while still relying heavily on hedging.

\begin{tcolorbox}[colback=gray!10, colframe=gray!60!black,boxrule=0.5pt,arc=3pt]
\textbf{Findings to RQ2:} AIACC-generated reports contain frequent hedge cues. Directive keywords appear more often in obligation statements and other content than in procedural instructions. These indicators reflect that there is still room for improvement in the report regarding determinacy and actionability.
\end{tcolorbox}
\section{Legal Coverage and Alignment (RQ3)}
\label{sec_rq2}
We evaluate the extent to which AIACCs cover the AI Act provisions. A checker that omits data governance, robustness, cybersecurity, or role-specific duties may redirect users away from safeguards the Act is intended to trigger. We therefore measure article coverage and examine selected legal alignment issues using the representative scenarios defined in Section~\ref{subsec_scenario_construction}. We do not establish definitive legal ground truth for every checker output, because the applicability of provisions depends on detailed facts, deployment context, and applicable exceptions. Our analysis characterizes coverage and alignment under the stated scenario assumptions rather than establishing the legal correctness.

\subsection{Evaluation Scope}
We use the article scope in~\cite{complai24}, which studies the technical interpretation of the AI Act, as a starting point for our evaluation scope. It focuses on requirements that can be translated into technical properties, such as robustness, cybersecurity, data governance, transparency, privacy, and documentation.
Following this rationale, we select 18 articles that can be examined in AIACCs' inputs and outputs. The scope spans AI literacy (Article 4), prohibited and high-risk classification (Articles 5--6), high-risk provider obligations (Articles 9--16), deployer obligations and fundamental rights (Articles 26--27), registration and transparency duties (Articles 49--50), and General-Purpose AI model obligations (Articles 51, 53, and 55).


\subsection{Article Coverage}
\textbf{Classification-Oriented Coverage.} Table~\ref{tab:article-new} indicates a highly uneven coverage pattern.
The most consistently covered provisions are prohibited practices (Article 5) and high-risk classification (Article 6), each covered by 10 of the 12 assessed checkers. Transparency obligations (Article 50) and GPAI systemic-risk classification (Article 51) are also common, each covered by 9 checkers. These provisions are comparatively easy to express as user-facing routing rules, where a checker can ask whether the system involves a prohibited practice, falls into a high-risk area, interacts with humans or generates content, or may qualify as a GPAI model with systemic risk. Their prominence is also consistent with the Act's risk-tiered structure and with the European Commission's early implementation efforts, such as issuing guidance on prohibited practices~\cite{ec2025prohibitedguidelines}, AI-system definition~\cite{ec2025aisystemdefinition}, and draft high-risk classification guidelines with practical examples~\cite{ec2026highriskguidelines}, all of which make classification more amenable to checker-style operationalization.

\textbf{Lifecycle Governance.} Coverage drops once the Act moves from classification to lifecycle governance. 
Articles 9--15 require risk management, data governance, technical documentation, record keeping, transparency for deployers, human oversight, and accuracy, robustness, and cybersecurity. These provisions are covered by only 4 or 5 checkers each, and they are almost entirely absent from dynamic branching questionnaires. This pattern suggests that many AIACCs stop at the point where compliance becomes evidence-intensive. Unlike classification, these provisions cannot be checked reliably through a small number of questions. They require information about organizational processes, datasets, logging, evaluation, human oversight design, cybersecurity controls, and post-deployment monitoring. Prior work also argues that AI Act compliance depends on auditable processes and evidence rather than one-off declarations~\cite{mokander2022conformity, golpayegani2024ai, marino2024compliancecardsautomatedeu}. The difficulty is particularly visible for record-keeping (Article 12) and accuracy, robustness, and cybersecurity (Article 15), each covered by only 4 checkers, where recent studies emphasize that legal requirements are hard to test, quantify, and translate into technical criteria~\cite{langer2025complexities, Nolte_2025}.

\textbf{GPAI Obligations.} GPAI coverage is more mixed. 
Systemic-risk classification (Article 51) is covered by 9 checkers, while obligations (Article 53 and 55) are covered by 7 and 6 checkers respectively. Checkers more readily identify whether a legal category may apply than assess the detailed duties that follow from that category. The European Commission has published guidelines for GPAI providers~\cite{ec2025gpaiguidelines} and the GPAI Code of Practice~\cite{ec2025gpaicode}, including chapters on transparency, copyright, safety, and security. These materials make GPAI obligations increasingly operationalizable, but many AIACCs still treat GPAI primarily as a classification question.

%
%
\begin{table}[t]
\centering
\caption{Coverage of EU AI Act Articles across AIACCs.
}
\label{tab:article-new}
\scriptsize
\setlength{\tabcolsep}{2.2pt}
\renewcommand{\arraystretch}{1.08}
\resizebox{\linewidth}{!}{%
\begin{tabular}{l|cccccc|cc|cc|cc}
\hline
\rowcolor{lightgray!15}
\textbf{Mode} & \multicolumn{6}{c|}{\textbf{DBQ}} & \multicolumn{2}{c|}{\textbf{FQ}} & \multicolumn{2}{c|}{\textbf{CA}} & \multicolumn{2}{c}{\textbf{TAT}} \\

\rowcolor{lightgray!85}
\textbf{(\#)} & \textbf{1} & \textbf{2} & \textbf{3} & \textbf{4} & \textbf{5} & \textbf{6} & \textbf{7} & \textbf{8} & \textbf{9} & \textbf{10} & \textbf{11} & \textbf{12} \\
\hline

\rowcolor{lightgray!50}
\multicolumn{13}{l}{\textbf{General}} \\

\rowcolor{lightgray!15}
Art. 4 (AI Literacy) & \cmark & \cmark & \xmark & \xmark & \xmark & \cmark  & \xmark & \xmark & \cmark & \cmark & \xmark & \cmark \\

\rowcolor{lightgray!50}
Art. 5 (Prohibited AI Practices) & \cmark & \cmark & \cmark & \cmark & \cmark & \cmark  & \cmark & \cmark & \cmark & \cmark & \xmark & \xmark \\

\hline
\rowcolor{lightgray!50}
\multicolumn{13}{l}{\textbf{High-Risk AI}} \\

\rowcolor{lightgray!15}
Art. 6 (High-Risk Classification) & \cmark & \cmark & \cmark & \cmark & \cmark & \cmark  & \cmark & \cmark & \cmark & \cmark & \xmark & \xmark \\

\rowcolor{lightgray!50}
Art. 9 (Risk Management System) & \xmark & \xmark & \xmark & \xmark & \xmark & \xmark & \cmark & \xmark & \cmark & \cmark & \cmark & \cmark \\

\rowcolor{lightgray!15}
Art. 10 (Data and Data Governance) & \xmark & \xmark & \xmark & \xmark & \xmark & \xmark & \cmark & \xmark & \cmark & \cmark & \cmark & \cmark \\

\rowcolor{lightgray!50}
Art. 11 (Technical Documentation) & \xmark & \xmark & \xmark & \xmark & \xmark & \xmark & \cmark & \xmark & \cmark & \cmark & \cmark & \cmark \\

\rowcolor{lightgray!15}
Art. 12 (Record-Keeping) & \xmark & \xmark & \xmark & \xmark & \xmark & \xmark & \cmark & \xmark & \cmark & \cmark & \xmark & \cmark \\

\rowcolor{lightgray!50}
Art. 13 (Transparency for Deployers) & \xmark & \xmark & \xmark & \xmark & \xmark & \xmark & \cmark & \xmark & \cmark & \cmark & \cmark & \cmark \\

\rowcolor{lightgray!15}
Art. 14 (Human Oversight) & \xmark & \xmark & \xmark & \xmark & \xmark & \xmark & \cmark & \xmark & \cmark & \cmark & \cmark & \cmark \\

\rowcolor{lightgray!50}
Art. 15 (Accuracy, Robustness, Cybersecurity) & \xmark & \xmark & \xmark & \xmark & \xmark & \xmark & \xmark & \xmark & \cmark & \cmark & \cmark & \cmark \\

\rowcolor{lightgray!15}
Art. 16 (Provider Obligations) & \cmark & \cmark & \xmark & \cmark & \xmark & \cmark & \xmark & \xmark & \cmark & \cmark & \xmark & \xmark \\

\rowcolor{lightgray!50}
Art. 26 (Deployer Obligations) & \cmark & \cmark & \xmark & \cmark & \xmark & \xmark & \xmark & \xmark & \cmark & \cmark & \xmark & \xmark \\

\rowcolor{lightgray!15}
Art. 27 (Fundamental Rights) & \cmark & \xmark & \xmark & \xmark & \xmark & \xmark & \xmark & \xmark & \cmark & \xmark & \cmark & \xmark \\

\rowcolor{lightgray!15}
Art. 49 (Registration) & \cmark & \xmark & \xmark & \xmark & \xmark & \cmark & \xmark & \xmark & \cmark & \xmark & \xmark & \xmark \\

\hline
\rowcolor{lightgray!50}
\multicolumn{13}{l}{\textbf{Transparency}} \\

\rowcolor{lightgray!50}
Art. 50 (Transparency Obligations) & \cmark & \cmark & \cmark & \cmark & \xmark & \cmark & \cmark & \xmark & \cmark & \cmark & \cmark & \xmark \\

\hline
\rowcolor{lightgray!50}
\multicolumn{13}{l}{\textbf{General-Purpose AI}} \\

\rowcolor{lightgray!15}
Art. 51 (GPAI Systemic Risk Classification) & \cmark & \cmark & \cmark & \cmark & \cmark & \cmark & \xmark & \cmark & \cmark & \xmark & \cmark & \xmark \\

\rowcolor{lightgray!50}
Art. 53 (GPAI Provider Obligations) & \cmark & \cmark & \xmark & \xmark & \xmark & \cmark & \xmark & \cmark & \cmark & \cmark & \cmark & \xmark \\

\rowcolor{lightgray!15}
Art. 55 (Systemic-Risk GPAI Obligations) & \cmark & \cmark & \xmark & \xmark & \xmark & \cmark & \cmark & \xmark & \cmark & \xmark & \cmark & \xmark \\

\hline
\end{tabular}
}
\end{table}

\textbf{Role-Specific Duties.}
Provider obligations (Article 16) are covered by 6 checkers and deployer obligations (Article 26) by 5, but fundamental-rights impact assessment (Article 27) and registration (Article 49) are covered by only 3 checkers each. Failing to distinguish provider, deployer, importer, distributor, or product-manufacturer responsibilities may give users a risk label without correct compliance guidance. This difficulty is also recognized in an AI supply chain study~\cite{marino2024compliancecardsautomatedeu}, which shows that AI Act responsibilities can be distributed across multiple actors and artifacts rather than located in a single system owner. Article 4 on AI literacy is covered by only half of the studied AIACCs. One reason could be that AI literacy is organizational rather than system-specific. It concerns whether providers and deployers ensure appropriate knowledge and training for persons dealing with AI systems. The European Commission has responded through Q\&A materials~\cite{ec2025ailiteracyQA} and a repository of AI literacy practices~\cite{ec2025ailiteracyRepo}. This illustrates a broader pattern, where system-specific routing questions are easier for AIACCs to cover, while provisions requiring organizational evidence are harder to encode.

\textbf{Checker Mode.} 
Checker mode affects article coverage. Questionnaires mainly prioritize provisions suitable for tree-like classification: most questionnaire tools cover Articles 5 and 6, but the six dynamic branching questionnaires do not cover Articles 9--15 except for limited role-specific references in Articles 16, 26, and 49. Conversational assistants appear broad because they can mention many provisions on request. Both conversational tools cover Articles 5, 6, 9--16, 26, and 50, although this coverage may depend on prompt wording and conversational trajectory. Technical analysis tools show the opposite profile. They omit Article 5 and Article 6 classification, but cover several lifecycle provisions in Articles 9--15 because their natural inputs are artifacts such as documentation or model outputs. Thus, modes do not simply differ in user interface and underlying mechanism but also embody different legal modeling. Questionnaires model compliance as classification, conversational assistants model it as explanation, and technical tools model it as evidence evaluation. This pattern should also be interpreted considering each checker's design goal. A risk-classification checker may mention Article 10 after identifying a high-risk system, but still not assess data governance practices in detail because its classification task is complete.


\begin{figure*}[t]
  \centering
  \includegraphics[width=.95\linewidth]{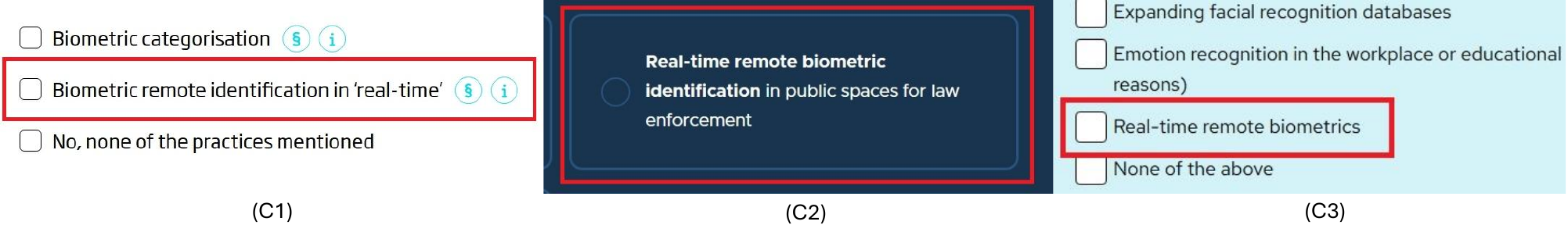}
  \caption{Different wordings of ``real-time biometric identification'' question across checkers C1-C3.
  }
  \label{fig_wording}
\end{figure*}

\subsection{Legal Alignment Observations}
During the article coverage analysis, we observe recurring patterns in terms of legal alignment. We present specific examples to unveil these patterns.

\textbf{Risk Misclassification.} The first alignment problem concerns the boundary between prohibited practices and high-risk systems. Prohibited-practice boundary logic can be fragile. An example error is to conflate real-time biometric identification with an Article 5 prohibition without distinguishing law-enforcement use from private deployment. For example, C3 requires users to select ``real-time biometric identification,'' but this choice immediately triggered a prohibited classification. To proceed further, users effectively had to choose ``none of the practices mentioned,'' which is legally inaccurate. C2 handles this more precisely by wording the question as ``real-time remote biometric identification in public places for law enforcement,'' thereby narrowing the trigger to the legally relevant context. Figure~\ref{fig_wording} shows how wording differences across C1, C2, and C3 affect whether biometric systems are treated. The underlying reason could be that some AIACCs encode prohibited practices as broad red flags without considering scope conditions and exceptions. This makes the checker easy to implement, but it collapses a structured legal test into a single yes or no trigger. The European Commission's guidance likewise emphasizes the need to interpret Article 5 in relation to the specific practice, actor, and context~\cite{ec2025prohibitedguidelines}. Checker questions that omit these qualifiers can therefore over-classify systems as prohibited or force legally inaccurate user answers.


\textbf{Ambiguous Duty Allocation.} Some AIACCs do not determine the role of the AIACC user in the AI supply chain. High-risk status imposes a multi-role obligation set. Providers may need to satisfy requirements such as technical documentation, quality management, conformity assessment and post-market monitoring, and deployers face obligations such as human oversight and monitoring. Prior work also points out that AI Act responsibilities may be distributed across multiple roles and artifacts~\cite{marino2024compliancecardsautomatedeu}. Therefore, role-insensitive outputs risk both \textit{over-inclusion} and \textit{under-inclusion}, where one user may receive duties they cannot perform, while another may miss their responsibilities. 

\begin{tcolorbox}[colback=gray!10, colframe=gray!60!black,boxrule=0.5pt,arc=3pt]
\textbf{Findings to RQ3:} Current checkers primarily operationalize the AI Act's risk classification layer. Coverage becomes sparse and mode-dependent for governance, documentation, role-specific, and security-relevant obligations. Article coverage is also affected by the mode and design goal of AIACCs.
\end{tcolorbox}

\section{Discussion}
\label{sec_discussion}

\subsection{Threat Sources}

Building on the aforementioned findings, we further identify three primary threat sources in the context of compliance checker adoption. 
The first threat is unreliable guidance under ordinary use. This resembles a similar problem in security and privacy compliance tooling, where users may over-rely on an automated signal even when the tool performs only partial or heuristic checking. An honest user may receive an incorrect, incomplete, inconsistent, or non-actionable assessment because the checker omits relevant obligations, oversimplifies conditions, or provides ambiguous natural language reasoning. When users treat a checker-generated report as evidence of compliance readiness, an inaccurate positive assessment may create a false sense of compliance~\cite{chung2024false,mazmudar2020mitigator,appg_paper}, allow relevant safeguards to remain unaddressed, and ultimately harm customers. 

The second threat is strategic or adversarial use. Users may learn how to frame inputs to obtain a more favorable output -- be it in the process of using an AIACC for determining the risk level of an AI system or checking its compliance with the AI Act requirements. A user can omit facts, choose ambiguous descriptions, or repeatedly interact with a conversational checker until the answer becomes more favorable. Consequently, the output from the AIACC may not accurately reflect the actual risk level or compliance status. For example, a high-risk AI system may be categorized as low risk, meaning that no further action is required, or the details in the generated report may not accurately reflect the system's actual characteristics. For LLM-based checkers, this threat also includes prompt manipulation and user-pleasing behavior, where the system changes its conclusion after pushback rather than preserving a stable legal rationale. As AIACCs become more user-friendly and the production of well-structured, professional-sounding compliance documentation becomes easier, there is a growing risk that (intentional) inaccuracies will be harder to detect during potential third-party audits. 

The third threat is privacy and confidentiality exposure through the checker itself. Compliance checking can require users to submit sensitive system descriptions, internal documentation, business plans, model behavior, or personal data processing details. This creates a conventional third-party service risk, where the tool that is supposed to support compliance may itself collect or process information whose storage, reuse or transfer is unclear. The risk may be amplified when AIACCs are operated by external providers or conversational platforms and when their data handling practices are not transparent.





\subsection{Design Barriers}

Most compliance checking is a decision-support process, where users provide information about an AI system, AIACCs map the information to legal context, and the output may influence subsequent safeguards.
We discuss the design barriers across studied AIACCs that may undermine the quality of the checking results.

\textbf{Lack of Input Validation.} 
AIACCs largely depend on user-provided descriptions but rarely validate whether those descriptions are complete, internally consistent, or legally meaningful. Among 12 AIACCs, we did not observe any explicit sanity check on user inputs before routing them to a legal conclusion. A user who misunderstands a term such as ``biometric identification'' can therefore be routed to the wrong result without being corrected. This creates a ``front-loaded'' or ``garbage in, garbage out'' failure mode. The checker appears automated, but much of the legal reasoning is implicitly delegated to the user at the input stage. Technical analysis tools partially avoid this problem by accepting artifacts such as model outputs or documentation, but they introduce another barrier that they assume users can prepare and interpret technical evidence. Thus, lightweight checkers are easy to use but fragile, while artifact-based tools are more evidence-oriented but inaccessible to many SMEs and individual developers.

\textbf{Over-simplified Legal Modeling.}
Many AIACCs compress legal reasoning into rigid forms. Fixed questionnaires minimize interaction cost, but their linear design forces heterogeneous AI systems through the same narrow path, where nuanced facts must be expressed as yes/no answers and outputs often appear as generic advice. Dynamic branching questionnaires are better aligned with the conditional structure of the AI Act but can still obscure why a specific branch was taken or which provision triggered a conclusion. Such simplification also happens in conversational assistants, where reasoning is conveyed through free-text chatting. Questionnaires generally lack mechanisms for detecting contradictory answers, testing alternative interpretations, or explaining edge cases. 

\textbf{Inherit Limitation, Consistency, and reproducibility.}
Conversational assistants address some usability barriers by allowing users to ask questions in natural language. This makes legal concepts easier to explore, but also introduces risks. LLM-based checkers inherit known limitations such as prompt injection~\cite{liu2023prompt}, hallucination~\cite{magesh2025hallucination}, and verbosity~\cite{zhang2024verbosity}.
In the compliance checking setting, these limitations are not merely cosmetic but can harm the checking validity. 
One example is shown in Figure~\ref{fig_model_incon}. The GPT-based checker C9 reached opposite conclusions for the same scenario. GPT-4o wrongly classified the AI system as prohibited, whereas GPT-5 correctly classified it as high-risk.
Conversational assistants are also difficult to reuse as compliance artifacts. Unlike structured questionnaires, they typically do not support systematic backtracking, branch comparison, or stable replay of a decision path. Long conversations may drift, and users may need to reconstruct earlier assumptions manually. Figure~\ref{fig_flow_incon} illustrates how a conversational flow can change under user pushback, while Figure~\ref{fig_model_incon} shows that the same scenario can produce materially different conclusions depending on the base model. This shows conversational assistants may be fragile for auditable compliance checking.

\textbf{Opaque Responsibility and Limited Reuse.} 
AIACCs often position their outputs as informational rather than authoritative, and many recommend additional professional review. 
This is understandable, but it creates a tension, where AIACCs lower the compliance burden but liability and interpretation remain with the user. In addition, users might have privacy concerns and hesitate to enter sensitive system descriptions, documentation, or business information, as AIACCs are not always transparent about data storage and reuse practices.
Outputs also vary in their downstream usefulness. AIACCs may provide transient web results or conversational text. However, compliance is not a one-time answer but requires documentation, review, revision, and communication across organizational roles. A result that cannot be exported and audited may help initial orientation but has limited ability to support sustained compliance work.

\textbf{Readability.} 
Legal reasoning does not guarantee usable guidance. As shown in Figure~\ref{fig_readability}, wording quality itself can become a barrier, where confusing double negatives, awkward question phrasing, or inconsistent confidence language make it harder for users to determine the meaning of guidance or checking output. This is especially problematic for AIACCs as users often lack legal expertise and rely on the checker's language.
Questionnaires may also introduce unnecessary interpretive burden through phrasing rather than legal substance. For example, C7 includes a double-negative yes/no question that is difficult to parse (Figure~\ref{fig_readability}a). C2's output includes phrases such as ``\textit{will likely}'' and ``\textit{also will need to}'', which mix different confidence levels and can make the guidance appear uncertain (Figure~\ref{fig_readability}b). These examples suggest that even when a checker covers the relevant obligations, unclear wording can still undermine user comprehension and trust.

\subsection{Design Principles}

Through the three RQs, we have identified systematic problems affecting AIACCs. Based on these observations, we derive the following design principles for compliance-support tools, in particular AIACCs, from our observations.

\textbf{Validate User Inputs.}
Many AIACCs depend heavily on user-provided information but rarely check whether the information is complete, internally consistent, or legally meaningful. This creates a front-loaded failure mode where low-quality user input may lead to incorrect results. Reliable AIACCs should, therefore, include input validation mechanisms, such as consistency checks and warnings when facts are insufficient for assessments. More generally, this also aligns with the evidence-oriented view of AI Act compliance, where claims about a system should be tied to evidence rather than unsupported self-descriptions~\cite{golpayegani2024ai,marino2024compliancecardsautomatedeu}.

\textbf{Make Outputs Actionable.}
AIACC-generated reports often identify legal themes without translating them into concrete next steps. While reports contain many directive words, our analysis indicates that these instructions often lack a concrete actor, object, or action. Reliable AIACCs should, therefore, express obligations as implementable tasks, and act as evidence-aware compliance workbenches.

\textbf{Support Reproducibility and Auditability.}
Compliance tools can be flexible but difficult to reuse as compliance artifacts. For example, conversations may drift, users may push the system toward different answers, and the same scenario can produce materially different conclusions. 
Reliable AIACCs should, therefore, record the input facts, assumptions, decision path, and generated output in an exportable format. Users should be able to replay or compare assessments after facts change or when laws evolve. This principle follows from auditability requirements in AI governance and conformity assessment~\cite{mokander2022conformity}. Prior privacy compliance studies~\cite{andow2019policylint,appg_paper} also leverage inspectable artifacts to detect contradictions or omissions.



\subsection{Recommendations for Stakeholders}
\textbf{Checker Providers.} Providers of AIACCs should be transparent about the checking logic behind results. They should design for input validation, actionable outputs, and support reproducibility and auditability. To accommodate revised regulatory requirements, providers could maintain a versioned legal knowledge base with explicit rules and source mappings, rather than revising checker design or fully relying on base model's responses. They should further avoid presenting orientation results as sufficient compliance outputs when downstream safeguards remain unassessed. 

\textbf{Organizations, SMEs, and developers.} Users should treat AIACCs as first-pass orientation tools, not as evidence of compliance. Our findings do not establish an overall best mode. Users should therefore select AIACCs by assessment task, not scenario alone. Questionnaires can support initial risk and role classification, conversational interfaces is user-friendly, and technical tools can examine documentation or evidence. None independently provides a complete compliance assessment. In particular, checker outputs should be reviewed against the actual system facts. When a resulting compliance report is vague or highly hedged, users should treat it as a prompt for further analysis rather than as a compliance artifact.

\textbf{Regulators and Standards Bodies.} Regulators should treat AIACCs as part of the AI Act implementation ecosystem. Public guidance should include machine-checkable examples. Standards bodies can also support this ecosystem by clarifying what evidence should be produced for obligations that are difficult to operationalize, such as human oversight and post-market monitoring.

\textbf{Researchers.} The research community should build empirical benchmarks and evaluation methods for AI Act compliance tooling. Our study provides an initial baseline of the landscape, but future work should examine longitudinal changes as the AI Act and checker implementations evolve. More broadly, automated compliance tools should be studied with the same scrutiny applied to other security- and privacy-relevant support tools, as their outputs can shape real-world engineering and governance decisions.

\section{Related Work}
\label{sec_related_work}

Prior work has examined how the EU AI Act can be translated from legal text into operational compliance practice. M{\"o}kander et al.~\cite{mokander2022conformity} frame conformity assessment and post-market monitoring as an AI auditing ecosystem. COMPL-AI maps AI Act requirements into measurable LLM benchmarks~\cite{complai24}. AIReg-Bench evaluates whether LLMs can assess AI Act compliance from technical documentation excerpts~\cite{marino2025aireg}. Several studies pinpoint  challenges in the context of robustness, cybersecurity~\cite{Nolte_2025}, human oversight~\cite{langer2025complexities} and standards~\cite{kilian2025european}. There are also efforts to build compliance-support artifacts, such as AI Cards~\cite{golpayegani2024ai} and Compliance Cards~\cite{marino2024compliancecardsautomatedeu} for automated AI Act analysis, AI Act technical documentation templates~\cite{lucaj2025techopstechnicaldocumentationtemplates}, and PaFRIA~\cite{ullstein2026proposing} for participatory fundamental-rights impact assessments.
Related empirical work has evaluated automated compliance tools, such as privacy-policy generators~\cite{appg_paper}. These studies reveal existing challenges. Our work studies a currently under-examined tool, e.g., AI Act compliance checkers. We empirically characterize existing AIACCs, analyze their input/output artifacts, and measure their legal coverage, thereby providing a baseline of the AIACC landscape.

\section{Concluding Remarks}
\label{sec_conclusion}

We present the first empirical study of publicly available EU AI Act compliance checkers. By characterizing 12 AIACCs, analyzing checker-generated reports, and evaluating legal coverage and alignment, we show that current AIACCs can lower the threshold for initial compliance. However, they remain early-stage orientation tools, rather than reliable compliance mechanisms. They vary in interaction mode, transparency, and user-friendliness. They often rely on oversimplified legal modeling and produce results characterized by hedging and weak actionability. Further, AIACCs cover risk classification provisions more consistently than lifecycle governance obligations. Our findings suggest that future compliance-support tools should be designed as evidence-aware and auditable tools that help users translate regulatory obligations into concrete safeguards for trustworthy AI deployment.

Compliance-support tools are becoming part of the practical infrastructure through which regulation is interpreted. However, the ease of access these tools provide can be misleading, as early errors may propagate through the compliance ecosystem. Incorrect risk labels and omitted role distinctions can influence what developers document, what safeguards managers fund, and what evidence organizations retain for later review. A low-quality compliance-support tool may give developers a false sense of confidence without evidence. From a legal perspective, our findings also highlight that the ecosystem of compliance-support tools is not itself standardized. AIACCs do not replace legal review, but they may shape decisions in early stages. We therefore argue that automated compliance tools should be treated as a research object and advocate more attention on such tools.

\bibliographystyle{IEEEtran}
\bibliography{References}

\section{Appendix}


\subsection{Definitions and Annotation Criteria of Characterization Items}
\label{app_def_c}
\textbf{Risk Classification} captures whether the checker assigns or helps determine the AI Act risk level of an AI system, such as prohibited, high-risk, limited-risk, minimal-risk, GPAI, or out of scope. This is the entry point for many downstream obligations. We coded \CIRCLE{} when the checker explicitly provides a risk classification, \LEFTcircle{} when risk classification is not returned as a clear assessment, and \Circle{} when the checker does not provide a risk-level determination.

\textbf{Article Mapping} captures whether the checker links its questions, outputs, or recommendations to specific AI Act articles or annexes. Article mapping supports legal traceability and helps users verify why a requirement applies. We coded \CIRCLE{} when specific articles are consistently linked to questions or results, \LEFTcircle{} when legal provisions are cited only selectively or generally, and \Circle{} when the checker gives no article-level traceability.

\textbf{User Guidance} captures whether the checker provides explanations, help text, definitions, tooltips, or other guidance that helps non-experts answer questions correctly. This is important because misunderstood legal terms can lead to incorrect risk classification or missing safeguards. We coded \CIRCLE{} when guidance is available throughout the assessment, \LEFTcircle{} when guidance appears only in very little cases, and \Circle{} when users receive no help.

\textbf{Examples} captures whether the checker provides concrete examples of AI systems, use cases, or answers. Examples can reduce ambiguity for users who are unfamiliar with legal categories such as biometric identification, deployer, or provider. We coded \CIRCLE{} when examples are provided systematically for relevant inputs or legal categories, \LEFTcircle{} when examples appear only occasionally, and \Circle{} when examples are absent.

\textbf{Interactive Q\&A} captures whether the checker supports follow-up questions or clarification during the assessment. We coded \CIRCLE{} when users can ask open-ended questions and receive responsive answers, \LEFTcircle{} when interaction is limited to scripted follow-up, and \Circle{} when the checker only supports fixed form navigation.

\textbf{Data Handling} captures whether the checker explains how user-submitted information is collected, processed, stored, shared, or protected. This item is directly relevant to privacy and security as users may submit sensitive descriptions of AI systems, datasets, business processes, or technical documentation. We coded \CIRCLE{} when the checker or linked policy clearly explains data collection and handling for submitted information (e.g., in a privacy policy), \LEFTcircle{} when only partial or generic disclosure is available, and \Circle{} when no meaningful data-handling information is provided.

\textbf{Checking Logic} captures whether the checker makes its assessment logic inspectable, for example by explaining decision paths, exposing rules, linking questions to provisions, or publishing source code. Transparent checking logic supports reproducibility, auditability, and trust. We coded \CIRCLE{} when users can inspect the rules, decision paths, or logic behind the assessment, \LEFTcircle{} when the checker exposes partial reasoning such as brief explanations, and \Circle{} when the assessment operates largely as a black box.

\textbf{Result Export} captures whether the checker allows users to save or export the result as a report, PDF, structured file, or other reusable artifact. Exportability matters because compliance work often requires documentation, internal review, and later audit. We coded \CIRCLE{} when the tool provides an exportable result artifact, and \Circle{} when no exportable result is available.

\subsection{Definitions and Annotation Criteria of Input and Output Items}
\label{app_def_io}
\textbf{System Description} captures whether the checker asks users to name or briefly describe the AI system. We coded \LEFTcircle{} when description can be accepted but only as optional or open-ended context.

\textbf{User's Contact Information} captures whether the checker asks for identifying or contact information, such as name, email, or organization. We coded \LEFTcircle{} when it is optional or requested only for follow-up.

\textbf{AI Model or System} captures whether the checker distinguishes between an AI model and a deployed AI system. This distinction matters under the AI Act because GPAI model obligations and AI-system obligations may attach to different actors and artifacts. We coded \LEFTcircle{} when description can be accepted but only as optional or open-ended context.

\textbf{Domain of Application} captures whether the checker asks where the AI system is used, such as employment, education, healthcare, law enforcement, or customer service.

\textbf{Intended Purpose} captures whether the checker asks what the AI system is intended to do. Intended purpose is legally important because many AI Act obligations depend on the provider's stated purpose and the deployer's actual use.

\textbf{Interaction Type} captures whether the checker asks how the AI system interacts with users or affected persons, such as chatbot interaction, decision support, content generation, or backend filtering. Interaction type helps determine transparency duties and potential user-facing risks.

\textbf{Biometric/Sensitive Functions} captures whether the checker asks about biometric identification, biometric categorization, emotion recognition, profiling, vulnerable groups, or other sensitive functions. These inputs are important because they may trigger prohibited-practice rules, high-risk classification, or privacy-relevant safeguards.

\textbf{User Role} captures whether the checker asks the actor or role of the user. Role identification is necessary as the AI Act allocates different obligations to different actors.

\textbf{Geographic Scope} captures whether the checker asks whether the system is placed on the EU market, used in the EU, or affects persons in the EU. 

\textbf{Documentation} captures whether the checker asks users to provide or upload documentation or similar evidence. This input can support stronger compliance assessment but may also raise confidentiality and data-protection concerns.

\textbf{Resulting Report} captures whether the checker generates report-like  structured result. We coded \LEFTcircle{} when the result is conversational or not structured.

\textbf{Benchmark Score} captures whether the checker generates compliance score based on the benchmarking results.

\textbf{Visualization} captures whether the checker provides visualization of materials such as knowledge graphs of legal text and model documentation.

\subsection{Report Analysis}
Detailed results of hedge sentence rate and keyword loss of specificity calculation are shown in Table~\ref{tab:hedge-sentence-rate} and Table~\ref{tab:keyword-loss-instructions-only} respectively.



\begin{table}[t]
\centering
\caption{Hedge Sentence Rate ($\downarrow$) by scenario and checker.}
\label{tab:hedge-sentence-rate}
\resizebox{0.95\linewidth}{!}{%
\begin{tabular}{lcccccccc|c}
\toprule
Scenario & C1 & C2 & C3 & C4 & C5 & C6 & C7 & C8 & Avg. \\
\midrule
S1 & 0.333 & 0.667 & 0.333 & 1.000 & 1.000 & 1.000 & 0.690 & 0.342 & 0.671 \\
S2 & 0.500 & 0.800 & 0.200 & 0.750 & 0.300 & 0.417 & 0.655 & 0.333 & 0.494 \\
S3 & 0.833 & 0.750 & 0.429 & 0.750 & 0.571 & 0.941 & 0.737 & 0.409 & 0.678 \\
S4 & 0.522 & 0.800 & 0.143 & 1.000 & 0.600 & 0.611 & 0.680 & 0.395 & 0.594 \\
S5 & 1.000 & 0.750 & 0.429 & 1.000 & 0.571 & 1.000 & 0.680 & 0.409 & 0.730 \\
S6 & 0.333 & 0.750 & 0.500 & 1.000 & 1.000 & 1.000 & 0.719 & 0.571 & 0.734 \\
\midrule
Avg. & 0.587 & 0.753 & 0.339 & 0.917 & 0.674 & 0.828 & 0.693 & 0.410 & 0.650 \\
\bottomrule
\end{tabular}%
}
\end{table}

\begin{table}[t]
\centering
\caption{Keyword Loss of Specificity ($\downarrow$) by scenario and checker. We report ``NA'' for the report that has no keyword-containing sentences.}
\label{tab:keyword-loss-instructions-only}
\resizebox{0.95\linewidth}{!}{%
\begin{tabular}{lcccccccc|c}
\toprule
Scenario & C1 & C2 & C3 & C4 & C5 & C6 & C7 & C8 & Avg. \\
\midrule
S1 & NA & NA & NA & NA & 0.000 & 1.000 & 0.563 & 0.714 & 0.569 \\
S2 & 0.833 & 1.000 & NA & 1.000 & 0.250 & 1.000 & 0.588 & 0.750 & 0.775 \\
S3 & 1.000 & 1.000 & 1.000 & 1.000 & 1.000 & 1.000 & 0.700 & 0.750 & 0.931 \\
S4 & 1.000 & 1.000 & NA & 1.000 & 1.000 & 1.000 & 0.600 & 0.800 & 0.914 \\
S5 & 1.000 & 1.000 & 1.000 & 1.000 & 0.000 & 1.000 & 0.600 & 0.750 & 0.794 \\
S6 & NA & NA & 1.000 & 1.000 & 1.000 & 1.000 & 0.700 & 1.000 & 0.950 \\
\midrule
Avg. & 0.958 & 1.000 & 1.000 & 1.000 & 0.542 & 1.000 & 0.625 & 0.794 & 0.840 \\
\bottomrule
\end{tabular}%
}
\end{table}

\end{document}